%% file: neurips_2026_camera_ready.tex
\documentclass{article}
\usepackage{graphicx}
\usepackage{float}
\usepackage{multirow}
\usepackage{wrapfig}

\usepackage{xcolor}
\newcommand{\rev}[1]{\textcolor{black}{#1}}

\PassOptionsToPackage{numbers,compress}{natbib}
\usepackage[final,main]{neurips_2026}
\setcitestyle{square,numbers}

\usepackage[utf8]{inputenc} 
\usepackage[T1]{fontenc}    
\usepackage{hyperref}       %

\usepackage{url}            
\usepackage{booktabs}       
\usepackage{amsmath}
\usepackage{amsfonts}     
\usepackage{nicefrac}       
\usepackage{microtype}      
\usepackage{xcolor}         
\usepackage{enumitem}
\usepackage{pifont}
\newcommand{\cmark}{\ding{51}}
\newcommand{\xmark}{\ding{55}}
\title{UltraVoxelGS: Voxel-First Feed-Forward Gaussian Splatting for 3D Ultrasound Reconstruction}

\author{%
  Yue Wang, Haoxin Chen, Xiaoqing Guo \\
  Department of Computer Science, Hong Kong Baptist University \\
  Hong Kong SAR, China \\
  \texttt{25481983@life.hkbu.edu.hk, 25482661@life.hkbu.edu.hk} \\
  \texttt{xiaoqingguo@hkbu.edu.hk}
}

\begin{document}

\maketitle

\begin{abstract}

We present \textbf{UltraVoxelGS}, a 3D ultrasound reconstruction method that achieves both \textbf{\textit{fast reconstruction}} and physically \textbf{\textit{faithful rendering}} for the first time, bringing live spatial feedback during 2D ultrasound scanning within reach. This is made possible by a feed-forward design that predicts a complete Gaussian scene representation from posed ultrasound images directly, entirely bypassing the per-scene optimization bottleneck that has confined prior methods to offline use. Realising such feed-forward prediction in this setting is non-trivial: unlike natural images, ultrasound slices share at most one-dimensional intersections and exhibit view-dependent appearance due to acoustic propagation and attenuation, violating the dense overlap and photometric consistency assumed by existing feed-forward Gaussian methods. Hence, to achieve fast reconstruction, we introduce a \textbf{\textit{feed-forward voxel-to-Gaussian pass}} that decouples representation size from input count by lifting all posed slices into a fixed-size voxel support and predicting Gaussian parameters directly. For faithful rendering, we propose \textbf{\textit{ultrasound-aware appearance rendering}} strategy factorize reflection from cumulative attenuation during training and exploit appearance continuity among spatially adjacent slices for efficient Render-time Appearance Adaptation. On both real-world  and simulated datasets, UltraVoxelGS achieves best PSNR/SSIM compared with per-scene optimization baselines while reducing inference from 10-20 minutes to approximately 10 seconds—over 35× faster—and scales to sequences exceeding 1,000 frames. Code is available at https://github.com/YuebyYue/UltraVoxelGS-camera-ready.

\end{abstract}

\section{Introduction}

\begin{figure}[t]
\centering
\includegraphics[width=\linewidth]{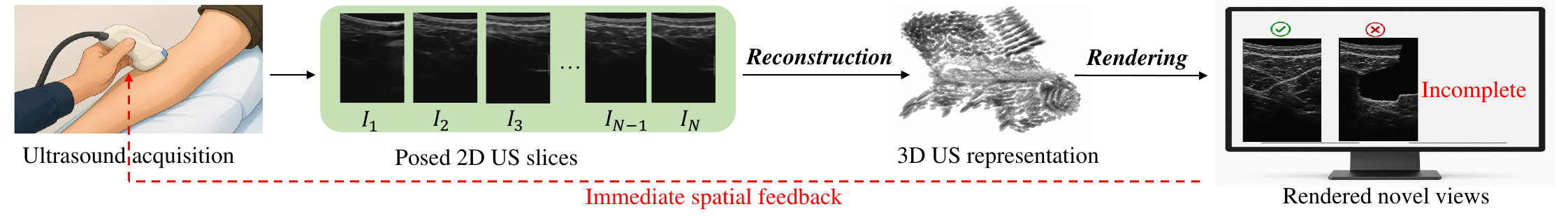}
\caption{\textbf{Clinical workflow.} A sequence of posed 2D ultrasound slices are \textit{reconstructed} into a 3D representation, from which novel views are \textit{rendered} to guide the operator toward unseen regions.}
\label{fig:runtime_workflow_pareto}
\end{figure}

Despite being the most widely used diagnostic imaging modality~\cite{fenster2001three,benacerraf2002three}, ultrasound remains fundamentally limited by its 2D acquisition: sonographers scan a sequence of slices and must mentally reconstruct the underlying 3D anatomy to determine which regions remain unseen. Incomplete coverage is therefore common, especially among less experienced sonographers, and often goes unnoticed during scanning. \rev{While dedicated 3D probes exist, their limited field of view makes them inadequate for large-anatomy or extended-field-of-view examinations, leaving 2D freehand scanning the dominant acquisition mode in practice.} These limitations motivate the need for \textit{real-time automatic 3D reconstruction from 2D ultrasound sequences}\cite{adriaans2024trackerless,shi2026real,guo2020sensorless}: during acquisition, a live 3D view can provide immediate spatial feedback to guide operators toward uncovered regions. After acquisition, a compact 3D representation also facilitates subsequent diagnosis and efficient archival. Realising this vision (Fig.~\ref{fig:runtime_workflow_pareto}) requires both \emph{\textbf{fast reconstruction}}—consolidating posed 2D slices into a compact 3D representation in seconds, not minutes—and \emph{\textbf{faithful rendering}}—synthesizing novel ultrasound views that respect the directional, cumulative-attenuation physics of acoustic imaging.

No existing approach satisfies both requirements. Classical interpolation-based methods~\cite{solberg2007freehand,chen2014reconstruction} are computationally cheap but treat ultrasound frames as generic intensity images, ignoring the directional propagation and cumulative attenuation of acoustic waves; they reproduce plausible intensities along scanned trajectories yet produce artifacts at novel viewpoints. Neural-field methods~\cite{yeung2021implicitvol,eid2025rapidvol,wysocki2024ultra,grutman2025implicit} achieve higher fidelity by learning a continuous scene representation, yet they rely on iterative per-scene optimization coupled with expensive volume rendering, making real-time reconstruction impractical. More recently, Gaussian splatting (GS)\cite{kerbl20233d} has emerged as a compelling alternative: its explicit primitives support efficient rasterization-based rendering, and recent adaptations~\cite{eid2025ultragauss} show that the formulation can accommodate ultrasound imaging assumptions. However, these 3D GS methods still reconstruct each scene through iterative optimization, meaning that they enable fast rendering but not fast reconstruction. \textit{The missing piece is not simply a faster renderer, but a feed-forward pathway from input slice sequences to a high-fidelity Gaussian representation.}

\begin{figure}[t]
  \centering
  \includegraphics[width=\linewidth]{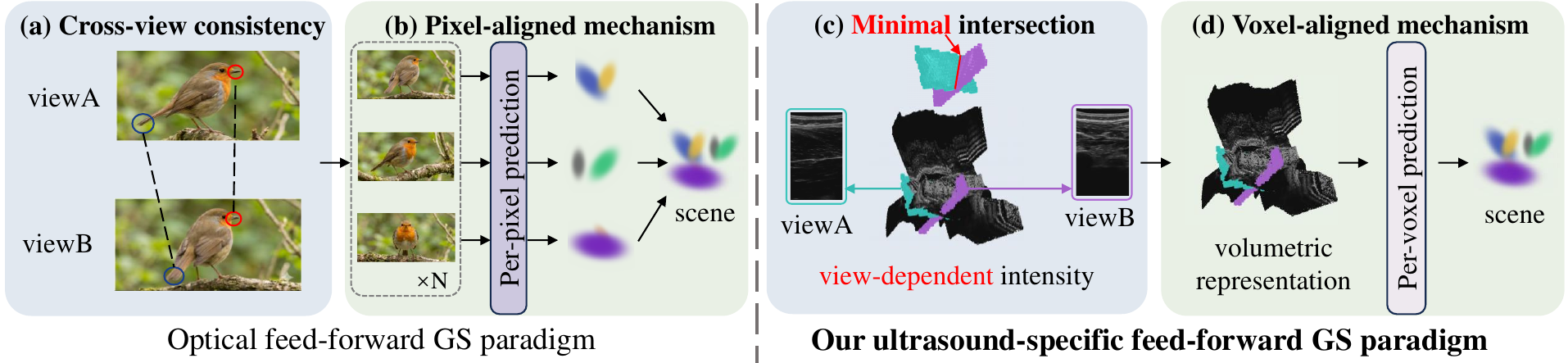}
  \caption{
\textbf{\textit{Optical feed-forward GS paradigm}}: 
{(a)} Natural images provide dense cross-view overlap and photometric consistency. 
{(b)} Existing methods predict pixel-aligned primitives from individual views.
\textbf{\textit{Our ultrasound-specific feed-forward GS paradigm}}: 
{(c)} Ultrasound slices share minimal intersection and exhibit view-dependent intensity.
{(d)} We lift posed slices into a unified voxel support and predict Gaussians from the shared voxel support.
}
  \label{fig:paradigm_shift}
\end{figure}

Feed-forward Gaussian prediction has been extensively explored in natural imaging~\cite{charatan2024pixelsplat,chen2024mvsplat,jiang2025anysplat,ye2024no,xu2025depthsplat}, but those methods assume dense view overlap and photometric consistency across viewpoints (Fig.~\ref{fig:paradigm_shift} (a)). Freehand ultrasound violates both assumptions: each frame captures only a thin 2D slice of the anatomy, so any two frames share at most a one-dimensional intersection (highlighted in Fig.~\ref{fig:paradigm_shift} (c)), providing far too little overlap for dense cross-view matching; and acoustic anisotropy makes the appearance of the same tissue view-dependent, undermining photometric consistency even along that shared line. The central challenge is therefore clear: each ultrasound slice carries only a thin cross-section of the anatomy, so the information from any single frame or even a small group of frames is far too sparse to support reliable Gaussian prediction. Existing multi-view feed-forward methods partially address this by jointly processing several frames (Fig.~\ref{fig:paradigm_shift} (b)), but their computational cost scales with the number of input views, making them impractical for the thousands of slices typical in freehand ultrasound. A scalable solution must therefore decouple representation size from input count (shown in Fig.~\ref{fig:paradigm_shift} (d)): \textit{first consolidate all posed slices into a fixed-size voxel support, then predict Gaussian parameters from this unified spatial context.}

We instantiate this idea in \textbf{UltraVoxelGS}, a feed-forward Gaussian splatting framework that, for the first time, enables both real-time reconstruction and physically faithful rendering for 3D ultrasound. 
Unlike prior feed-forward GS methods that generate pixel-aligned Gaussians from individual views and aggregate them afterward, we use probe poses to lift 2D ultrasound slices into 3D space and interpolate them into a unified voxel support, consolidating sparse observations into a coherent voxel support. We then predict Gaussian parameters directly from this shared voxel support, turning reconstruction into \textit{\textbf{a feed-forward voxel-to-Gaussian pass}}, entirely avoiding long per-scene optimization. 
For rendering, a separate challenge arises from the physics of ultrasound image formation: ultrasound pixel intensity couples local reflection/scattering with cumulative attenuation, making appearance strongly direction-dependent, yet this coupling also means that slices acquired under nearby probe poses can share similar local appearance. We exploit both aspects through the proposed \textit{\textbf{ultrasound-aware appearance rendering}} strategy, which decouples reflection/scattering from attenuation during training and leverages the local appearance similarity among neighboring slices for lightweight Render-time Appearance Adaptation. 
We demonstrate for the first time that feed-forward Gaussian splatting can match the reconstruction quality of per-scene optimization methods in ultrasound while reducing inference time from \texttt{10-20} minutes to \texttt{10} seconds, bringing 3D ultrasound reconstruction within reach of real-time clinical use. 

In summary, our contributions are fourfold:
\begin{itemize}[leftmargin=*, itemsep=2pt, parsep=0pt]
\vspace{-0.2cm}

\item We propose {UltraVoxelGS}, the first feed-forward Gaussian splatting framework for 3D ultrasound, enabling both real-time reconstruction and physically faithful rendering.

\item We design a \textit{{feed-forward voxel-to-Gaussian pass}} that lifts posed slices into a unified voxel support and predicts Gaussian parameters directly, achieving real-time 3D reconstruction.

\item We introduce \textit{{ultrasound-aware appearance rendering}} that decouples reflection/scattering from attenuation and exploits local appearance continuity for lightweight Render-time Appearance Adaptation, providing high-fidelity rendering at sub-second latency for operator inspection.

\item We validate on two datasets spanning diverse anatomies, matching per-scene optimization baselines in quality with over $35\times$ speedup and scaling to sequences exceeding $1{,}000$ frames.

\end{itemize}

\section{Related Works}

\textbf{3D Ultrasound Reconstruction}
Traditional methods depend on interpolation to aggregate images, such as kernel-based interpolation~\cite{chen2014reconstruction} to create an isotropic 3D volume. ImplicitVol~\cite{yeung2021implicitvol,chen2024neural,song2025implicitcell} and RapidVol~\cite{eid2025rapidvol} extend this interpolation from discrete representation to continuous representation. However, all of these construct an isotropic volume, violating the anisotropic property of ultrasound. The emergence of NeRF-based methods, such as Ultra-NeRF~\cite{wysocki2024ultra,dagli2024nerf}, formulate ultrasound parameters as continuous parameter field, then rendering images according to ultrasound imaging, following anisotropic property. These methods achieve high-quality synthesis, but suffer from low rendering speed coupled with dense neural field querying. As an alternative, GS methods model the scene as a set of Gaussian primitives, achieving fast rendering by rasterization. Recent methods focus on ultrasound imaging assumptions, assuming the imaging plane as a slice of the scene~\cite{eid2025ultragauss, guo2026visually,krahenmann2025slices,kong2026gaussianpile} or adopting a variable frustum projection~\cite{yang2025ultrags}. These methods render high-quality images efficiently in specific settings, but still belong to per-scene optimization.

\textbf{Feed-Forward 3D Reconstruction}
Feed-forward methods~\cite{charatan2024pixelsplat} have emerged in natural imaging. These methods predict Gaussians in a single feed-forward pass. Most existing works focus on a pixel-aligned mechanism, directly predicting per-pixel Gaussian primitives and then aggregating them. More recent extensions improve scalability through structured aggregation or voxel-aligned prediction~\cite{jiang2025anysplat,wang2025volsplat}. Similar feed-forward ideas have also been explored in projection-based medical imaging such as sparse-view CT~\cite{liu2025x}.  These methods rely on stronger cross-view correspondence but each observation in ultrasound is only a thin probe-aligned slice, cross-view overlap is extremely limited, and appearance is strongly view-dependent due to acoustic anisotropy. Therefore, existing feed-forward pipelines cannot be directly applied to freehand ultrasound.

\section{Methodology}

\subsection{Problem Reformulation}

Before presenting UltraVoxelGS, we formalize the mismatch between optical feed-forward GS paradigm and freehand ultrasound reconstruction. We first review the pixel-aligned feed-forward GS formulation used in natural imaging, then characterize ultrasound frames as sparse posed slice observations. This motivates a voxel-first reformulation that consolidates all slices into a coherent 3D support before predicting Gaussian primitives.

\textbf{Pixel-aligned mechanism.}
Optical feed-forward GS paradigm formulates reconstruction as view-conditioned primitive prediction. Given a set of calibrated images $\{(I_i,T_i)\}_{i=1}^{N}$, a network predicts Gaussian primitives from images and aggregates them into a continuous scene representation:
\begin{equation}
\mathcal{G}
=
\mathcal{A}
\left(
\left\{
\phi_\theta(I_i,T_i)
\right\}_{i=1}^{N}
\right),
\label{eq:standard_form}
\end{equation}

where $\phi_\theta$ denotes the feed-forward Gaussian predictor, $\mathcal{A}$ denotes cross-view aggregation, and $\mathcal{G}$ is the Gaussian scene representation. This formulation implicitly assumes that different views provide sufficiently overlapping supports and approximately view-consistent appearance, so that primitives predicted in view-specific supports can be matched, fused, and regularized.

\textbf{Ultrasound slice observations.}
Freehand ultrasound has a fundamentally different observation geometry. Each frame is a thin 2D slice embedded in 3D space by its probe pose, rather than a perspective image with a wide viewing frustum. We denote a pixel on an ultrasound slice as $p=(u,z)$, where $u$ is the lateral coordinate and $z$ is the depth coordinate. For the $i$-th frame, the corresponding 3D position is
\begin{equation}
\label{eq:slice_lifting_prelim}
\mathbf{x}_i(p)
=
T_i [u,0,z,1]^\top ,
\qquad p=(u,z)\in\Omega_i ,
\end{equation}
where $\Omega_i$ is the image domain and $T_i$ maps the probe coordinate system to the world coordinate system. The observed slice support is therefore
\begin{equation}
\Pi_i =
\left\{
\mathbf{x}_i(p)
\mid
p\in\Omega_i
\right\},
\end{equation}
where $\Pi_i$ can be viewed as the 2D surface formed by all pixels of frame $i$ after pose lifting. Unlike a perspective image that covers a broad 3D viewing frustum, an ultrasound frame only contributes a thin slice of spatial evidence. For two frames $i$ and $j$, the supports $\Pi_i$ and $\Pi_j$ typically share at most a line-like intersection within the anatomy (refer to Fig.\ref{fig:paradigm_shift} (c)). Thus, ultrasound slices do not provide the broad shared support required by Eq.~\eqref{eq:standard_form}. Although probe poses lift slice pixels into 3D space, the lifted pixels remain irregular observations on sparse 2D slice surfaces rather than a complete voxel support. 
Constructing $V$ therefore requires consolidating these posed observations onto a canonical voxel grid, where many voxel locations are only weakly constrained by nearby slice evidence due to the thin and weakly overlapping imaging planes. These observations suggest that the bottleneck is the absence of an explicit shared support before prediction.

\textbf{Voxel-first Problem Reformulation.} Instead of predicting Gaussians directly from images, to explicitly establish a shared support, we first consolidate all posed ultrasound slices into a unified voxel support:
\begin{equation}
\{(I_i,T_i)\}_{i=1}^{N}
\longrightarrow
V
\longrightarrow
\mathcal{G},
\end{equation}
where $V\in\mathbb{R}^{X\times Y\times Z}$ denotes a canonical voxel grid over the reconstruction volume. 
After this consolidation, Gaussian prediction is driven by the voxel support rather than by all input image pixels, so its scale is controlled by the voxel resolution $XYZ$ instead of the raw observation count $N|\Omega|$.

\begin{figure}[t]
  \centering
  \includegraphics[width=0.82\linewidth]{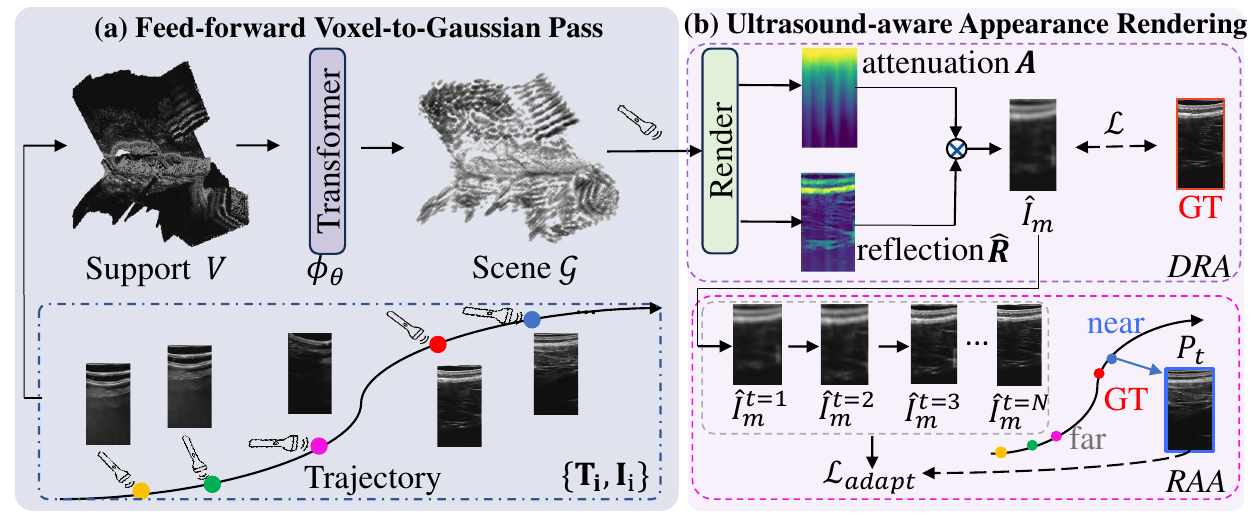}
\caption{
\textbf{Overview of UltraVoxelGS.}
\textbf{(a)} Posed ultrasound slices $\{T_i,I_i\}$ are lifted into a shared voxel support $V$, which is mapped by a feed-forward Transformer $\phi_\theta$ to the Gaussian scene representation $\mathcal{G}$.
\textbf{(b)} Ultrasound-aware appearance rendering uses Decoupled Reflection and Attenuation Rendering (DRA) to render each target slice as cumulative attenuation $A$ modulating anisotropic reflection $\hat R$ for image-level supervision. At render time, Render-time Appearance Adaptation (RAA) uses a nearby observed slice $P_t$ as an appearance anchor to refine target-view appearance.
}
  \label{fig:pipeline}
\end{figure}

We instantiate this reformulation in \textbf{UltraVoxelGS}. As shown in the overall pipeline (Fig.~\ref{fig:pipeline}), we first lift posed slices $\{(I_i,T_i)\}_{i=1}^{N}$ into a unified voxel support~$V$, then predict Gaussian primitives $\mathcal{G}$ from local voxel regions via a feed-forward voxel-to-Gaussian pass (Sec.~\ref{sec:forward}), and render novel slices through the ultrasound-aware appearance rendering strategy (Sec.~\ref{sec:rendering}).


\subsection{Feed-forward Voxel-to-Gaussian Pass}\label{sec:forward}

Our first step is to lift posed 2D slices $\{(I_i,T_i)\}_{i=1}^{N}$ into a unified voxel support~$V$. Following Eq.~\ref{eq:slice_lifting_prelim}, each pixel $p=(u,z)$ in frame $I_i$ is lifted to its 3D world position $\mathbf{x}_i(p)$, producing a set of irregularly distributed 3D observations$\mathcal{P}=\{(\mathbf{x}_i(p),I_i(p))\mid p\in\Omega_i,\ i=1,\ldots,N\}$, where $\mathbf{x}_i(p)$ denotes location and $I_i(p)$ is the corresponding pixel value. We then construct $V$ over the reconstruction volume. Let $\mathcal{X}_V$ denote the set of voxel centers. For a voxel center $\mathbf{x}\in\mathcal{X}_V$, we define $\mathcal{N}_k(\mathbf{x})$ as the up to $k$ nearest lifted observations within a radius $\alpha$:
\begin{equation}
\mathcal{N}_k(\mathbf{x})
\subset
\left\{
(i,p)
\mid
\|\mathbf{x}-\mathbf{x}_i(p)\|_2 \le \alpha
\right\}.
\end{equation}
The voxel intensity is computed by local inverse-distance interpolation:
\begin{equation}
\label{eq:voxel_interpolation}
V(\mathbf{x})
=
\frac{
\sum_{(i,p)\in\mathcal{N}_k(\mathbf{x})}
\omega_{i,p}(\mathbf{x}) I_i(p)
}{
\sum_{(i,p)\in\mathcal{N}_k(\mathbf{x})}
\omega_{i,p}(\mathbf{x}) + \epsilon
},
\qquad
\mathbf{x}\in\mathcal{X}_V,
\end{equation}
where $\omega_{i,p}(\mathbf{x})=(\|\mathbf{x}-\mathbf{x}_i(p)\|_2+\delta)^{-1}$, and $\delta$ and $\epsilon$ are small constants for numerical stability. Restricting interpolation to a local top-$k$ neighborhood avoids filling voxels from distant observations and limits the spread of unreliable measurements from  shadowed regions. The resulting voxel support provides a shared voxel support for subsequent Gaussian inference.

However, for dense ultrasound sequences, the voxel support can remain high-resolution, making it computationally prohibitive to predict Gaussian primitives over the entire volume. Fortunately, doing so is also unnecessary: each frame only observes a thin imaging slice and its nearby acoustic slab, while voxels far from this slab contribute little to the rendering of that frame.
We refer to this property as \textit{thin-slab locality}, and use it to restrict voxel-to-Gaussian prediction to local sub-volumes. For each target frame, we extract a local block $\mathcal{B}_m$ from the consolidated voxel support that covers the imaging slice and its nearby acoustic slab, and predict Gaussian primitives only within this block. Because this block is carved from the full voxel support—which has already fused posed observations across the entire input sequence—prediction operates on dense, multi-view spatial context rather than sparse single-frame information. 

Following this principle, for a supervised frame or target pose $m$, let $\mathcal{B}_m \subset \mathcal{X}_V$ denote the view-interacted local region associated with its imaging slice and nearby acoustic slab.
We tokenize $\mathcal{B}_m$ into regular voxel patches $\{P_j\}_{j=1}^{J_m}$, where $P_j\in\mathbb{R}^{n\times n\times n}$.
A Transformer-based predictor $\Phi_\theta$ operates on these voxel tokens and predicts local voxel-aligned Gaussian primitives:
\begin{equation}
\label{eq:ff_gs}
\{\mathcal{G}_j\}_{j=1}^{J_m}
=
\Phi_\theta(\{P_j\}_{j=1}^{J_m}),
\qquad
\widetilde{\mathcal{G}}_m
=
\bigcup_{j=1}^{J_m}
\mathcal{G}_j .
\end{equation}
Each predicted Gaussian primitive is parameterized as
$g=\{\mathbf{c}_g,\Sigma_g,\boldsymbol{\rho}_g,\beta_g\}$,
where $\mathbf{c}_g$ denotes the voxel-aligned Gaussian center, $\Sigma_g$ denotes the covariance, $\boldsymbol{\rho}_g\in\mathbb{R}^{9}$ denotes second-order spherical-harmonic coefficients for anisotropic reflection, and $\beta_g$ denotes the attenuation coefficient.

Since different view-interacted local regions can overlap, the same spatial location may receive multiple predicted Gaussian parameter vectors.
Let $\mathbf{a}_k(\mathbf{x})$ denote the parameter vector predicted for location $\mathbf{x}$ by the $k$-th local prediction.
We use a unified fusion rule: duplicated predictions are averaged, while non-duplicated predictions correspond to $K_{\mathbf{x}}=1$ and are kept unchanged:
\begin{equation}
\label{eq:fusion_strategy}
\bar{\mathbf{a}}(\mathbf{x}) =
\frac{1}{K_{\mathbf{x}}}
\sum_{k=1}^{K_{\mathbf{x}}}
\mathbf{a}_k(\mathbf{x}).
\end{equation}

During inference, UltraVoxelGS constructs the voxel support $V$ once from the input sequence, predicts Gaussian primitives from the view-interacted local regions, and assembles them using Eq.~\ref{eq:fusion_strategy}.
The fused and non-duplicated predictions together form the Gaussian scene representation $\mathcal{G}$. Importantly, this local-to-global assembly is not merely a computational shortcut: each local prediction is tied to the acoustic slab relevant to a target slice, while its input region is selected from the consolidated voxel support that integrates posed observations from the sequence. 
UltraVoxelGS thus avoids redundant whole-support prediction while preserving the slice-relevant information needed for accurate ultrasound rendering, enabling scalable reconstruction for long ultrasound sequences.






\begin{wrapfigure}[14]{r}{0.56\linewidth}
\vspace{-0.8em}
\centering
\includegraphics[width=\linewidth]{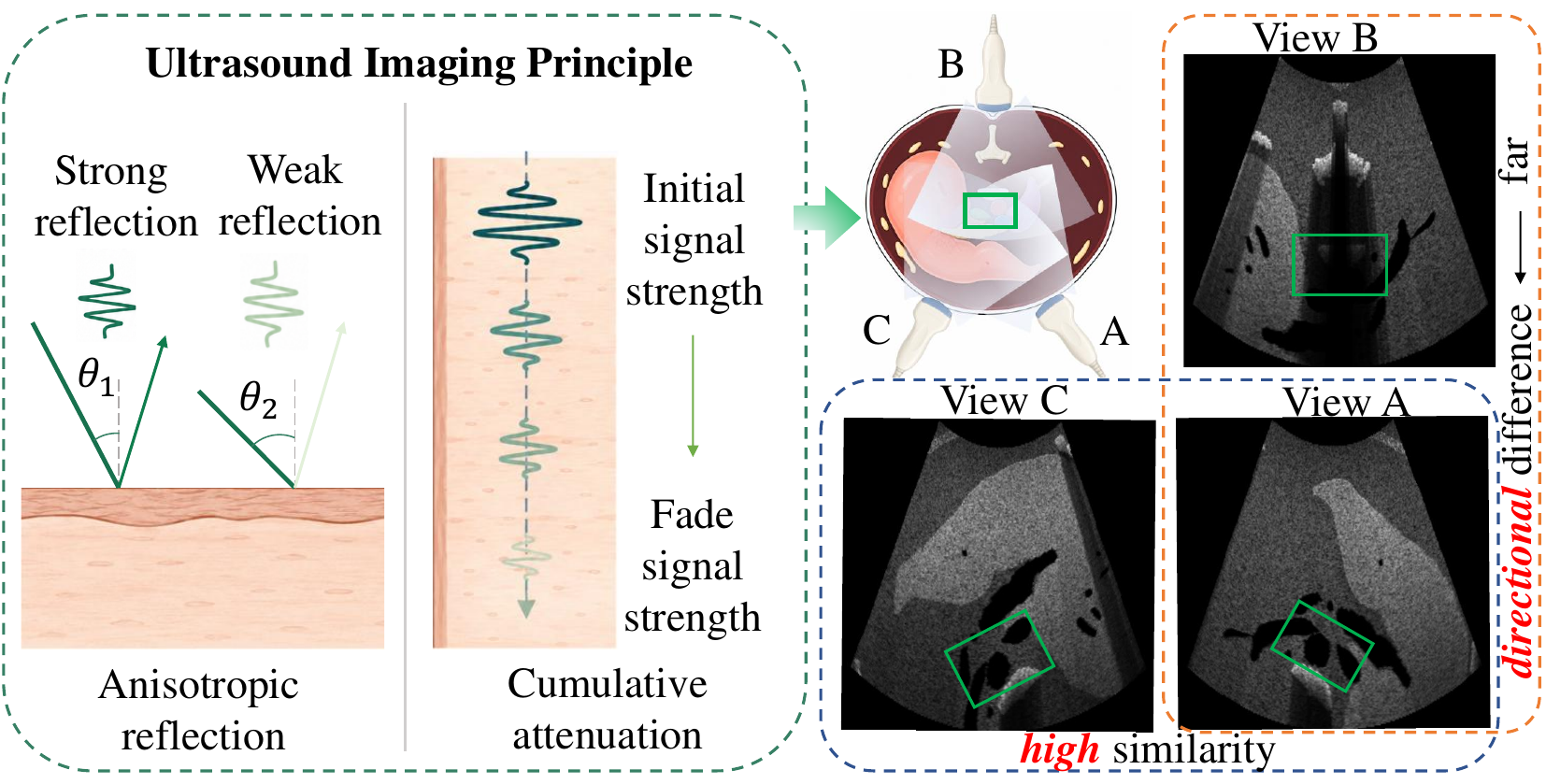}
\vspace{-0.6em}

\refstepcounter{figure}
\label{fig:imaging_theory}
\footnotesize
\noindent\textbf{Figure~\thefigure:}
\textbf{Ultrasound appearance modeling.}
Distant views are direction-dependent, while nearby views remain locally similar for Render-time Appearance Adaptation.

\end{wrapfigure}

\subsection{Ultrasound-aware Appearance Rendering}
\label{sec:rendering}

Given the predicted Gaussian scene representation $\mathcal{G}$, faithful rendering requires preserving ultrasound-specific image formation rather than treating the target slice as a generic intensity image. 
As shown in Fig.~\ref{fig:imaging_theory}, ultrasound appearance is governed by anisotropic reflection and cumulative attenuation\cite{jensen2002ultrasound}: even for the same anatomical plane, images can vary significantly with probe pose, as different poses induce different acoustic paths through the tissue. Only when probe poses are sufficiently close, and thus the acoustic paths are similar, do the resulting images exhibit similar appearance. 
Based on these properties, we introduce \textit{ultrasound-aware appearance rendering} strategy, consisting of \textit{Decoupled Reflection and Attenuation Rendering} for training supervision and \textit{Render-time Appearance Adaptation} for refinement from nearby observed slices.

\subsubsection{Decoupled Reflection and Attenuation Rendering}
\label{subsec:formulation}

The predicted Gaussian representation is rendered into ultrasound slices for supervision.
For a target frame $m$ and pixel $p=(u,z)\in\Omega_m$, we model ultrasound intensity as anisotropic reflection modulated by cumulative attenuation:
\begin{equation}
\label{eq:intensity}
I_m(p)
=
R_m(p,\mathbf{d}_m(p))
\cdot
\exp
\left(
-\int_{0}^{z}
\beta_m(u,\zeta)\,d\zeta
\right),
\end{equation}
where $\mathbf{d}_m(p)$ denotes the acoustic propagation direction under pose $T_m$, $R_m$ denotes the anisotropic reflection response, and $\beta_m$ denotes the attenuation coefficient.

Using the predicted Gaussian primitives, we instantiate this model with Decoupled Reflection and Attenuation Rendering.
Let $\mathcal{G}_m(p)$ be the set of Gaussians contributing to pixel $p$ under pose $T_m$, and let $w_{g,m}(p)$ be the contribution weight of Gaussian $g$.
We compute the anisotropic reflection term as
$\hat R_m(p)=\sum_{g\in\mathcal{G}_m(p)} S_{\mathrm{SH}}(\mathbf{d}_m(p);\boldsymbol{\rho}_g)\,w_{g,m}(p)$,
where $S_{\mathrm{SH}}(\mathbf{d}_m(p);\boldsymbol{\rho}_g)$ denotes the second-order spherical-harmonic response parameterized by $\boldsymbol{\rho}_g$.
The local attenuation coefficient is computed as
$\hat\beta_m(p)=\sum_{g\in\mathcal{G}_m(p)} \beta_g\,w_{g,m}(p)$,
and the accumulated transmittance along depth is
$A_m(p)=\exp(-\sum_{\zeta<z}\hat\beta_m(p_\zeta)\Delta z)$ with $p_\zeta=(u,\zeta)$.
The final rendered intensity is then $\hat I_m(p)=A_m(p)\hat R_m(p)$.
This renderer is slice-consistent and physically decoupled: $\hat R_m(p)$ captures anisotropic reflection, while $A_m(p)$ captures cumulative attenuation.

During training, the fused Gaussian representation $\mathcal{G}$ is optimized with a rendering loss and an overlap consistency loss:
\begin{equation}
\label{eq:total_loss}
\mathcal{L}
=
\sum_{m\in\mathcal{M}}
\sum_{p\in\Omega_m}
\left\|
\hat I_m(p)-I_m(p)
\right\|_2^2
+
\lambda
\sum_{\mathbf{x}\in\mathcal{X}_{\mathrm{ov}}}
\frac{1}{K_{\mathbf{x}}}
\sum_{k=1}^{K_{\mathbf{x}}}
\left\|
\mathbf{a}_k(\mathbf{x})-\bar{\mathbf{a}}(\mathbf{x})
\right\|_2^2,
\end{equation}
where $\mathcal{X}_{\mathrm{ov}}$ denotes spatial locations with multiple predictions. 
The first term supervises the final rendered intensity $\hat I_m(p)$ against the observed image $I_m(p)$, without requiring explicit reflection or attenuation annotations. 
The second term addresses a consistency issue: since thin-slab locality allows the same spatial location to be predicted from multiple overlapping slabs, the resulting Gaussian parameters may diverge across slabs. The overlap consistency loss penalizes the deviation of each predicted parameter vector $\mathbf{a}_k(\mathbf{x})$ from its fused average $\bar{\mathbf{a}}(\mathbf{x})$ over all $K_{\mathbf{x}}$ overlapping predictions at location $\mathbf{x}$, and $\lambda$ balances rendering fidelity and overlap consistency.

\subsubsection{Render-time Appearance Adaptation}
\label{subsec:tta}

The shared voxel support provides a common prediction domain, but voxel support interpolation can smooth local ultrasound appearance and reduce slice-level sharpness. 
Motivated by local appearance continuity, we use a nearby observed slice as an appearance anchor to provide sharper local appearance evidence for the target rendering. 
Given the initial Gaussian representation
$\mathcal{G}_0=\{(\mathbf{c}_g^0,\Sigma_g^0,\boldsymbol{\rho}_g^0,\beta_g^0)\}$ predicted by UltraVoxelGS, we freeze the geometric parameters $(\mathbf{c}_g^0,\Sigma_g^0)$ and update only a transient copy of the appearance parameters $(\boldsymbol{\rho}_g,\beta_g)$ for the target rendering. For a target pose $T_t$, we select the nearest observed source frame as the appearance anchor:

\begin{equation}
\label{eq:tta_nearest}
i^*(t)
=
\arg\min_{i\in\mathcal{S}}
\left\|
\mathbf{o}_i-\mathbf{o}_t
\right\|_2,
\qquad
P_t = I_{i^*(t)},
\end{equation}
where $\mathcal{S}$ denotes the set of observed frames, $\mathbf{o}_i$ and $\mathbf{o}_t$ denote the probe centers of poses $T_i$ and $T_t$, respectively, and $P_t$ is the selected source frame used as the appearance anchor. Then, the adaptation objective, denoted as $\mathcal{L}_{\mathrm{adapt}}$ in Fig.~\ref{fig:pipeline}(c), optimizes the transient appearance copy while keeping geometry fixed:
\begin{equation}
\label{eq:tta_opt}
(\boldsymbol{\rho}_t^*,\boldsymbol{\beta}_t^*)
=
\arg\min_{\boldsymbol{\rho},\boldsymbol{\beta}}
\underbrace{
\left\|
\mathcal{R}
\left(
\mathcal{G}
\left(
\mathbf{c}^0,\Sigma^0,\boldsymbol{\rho},\boldsymbol{\beta}
\right);
T_{i^*(t)}
\right)
-
P_t
\right\|_2^2
+
\gamma
\left\|
\boldsymbol{\rho}-\boldsymbol{\rho}^0
\right\|_2^2
+
\eta
\left\|
\boldsymbol{\beta}-\boldsymbol{\beta}^0
\right\|_2^2
}_{\mathcal{L}_{\mathrm{adapt}}}.
\end{equation}
where $\mathcal{R}(\mathcal{G};T)$ denotes the DRA renderer, and $\gamma,\eta$ regularize the adapted appearance parameters toward the feed-forward prediction. The first term uses the selected source frame $P_t$ as an appearance anchor, while the regularization terms keep the transient update close to the feed-forward appearance. The adapted parameters are used only to render the target pose $T_t$ and are discarded afterward. Because the adapted appearance parameters are discarded after rendering each target view, this stage only affects per-view rendering and leaves the reconstructed global representation unchanged.

\section{Experiments}

\subsection{Datasets and implementation details}

\textbf{Dataset A: Real-world Robotic Ultrasound Video Sequences.} This dataset consists of robot-assisted lumbar spine ultrasound scans with synchronized probe pose tracking 
~\cite{cavalcanti2025large}, acquired from 63 healthy volunteers \rev{(used under CC BY 4.0).} Each scan contains over 3K densely sampled frames. In the default setting, we uniformly sample every 5-th frame as input and supervision, and evaluate on frames with index $3 \bmod 5$ to avoid near-duplicate testing.

\textbf{Dataset B: Simulated Liver Ultrasound Sequences.} This dataset contains 100 simulated liver ultrasound sequences rendered from abdominal CT volumes from 3D-IRCADb~\cite{soler20103d} and CRLM-CT~\cite{simpson2024preoperative} based on LOTUS~\cite{velikova2023lotus} \rev{(used under CC BY-NC-ND 4.0 and CRLM-CT under CC BY 4.0).} We hold out every 7-th frame as novel views and use the remaining frames as input and supervision.

\rev{All datasets are used for non-commercial research purposes in accordance with their respective terms.} Additional implementation details, baseline settings, and dataset/tool license information are provided in Appendix~\ref{app:exp_details}.

\begin{table}[t]
\centering
\caption{{Quantitative comparison on real-world and simulated ultrasound datasets.} 
The best and second-best results are highlighted in bold and underline, respectively.}
\label{tab:main_results}
\resizebox{\linewidth}{!}{
\begin{tabular}{l|c|ccc|ccc}
\toprule
\multirow{2}{*}{Method} 
& \multirow{2}{*}{Type}
& \multicolumn{3}{c|}{Dataset A: Real-world} 
& \multicolumn{3}{c}{Dataset B: Simulated} \\
\cmidrule(lr){3-5}\cmidrule(lr){6-8}
& 
& PSNR$\uparrow$ & SSIM$\uparrow$ & LPIPS$\downarrow$
& PSNR$\uparrow$ & SSIM$\uparrow$ & LPIPS$\downarrow$ \\
\midrule

Kernel Interpolation~\cite{solberg2007freehand}
& Interpolation
& 13.52 & 0.3952 & 0.5419
& 17.47 & 0.2391 & 0.6259 \\

ImplicitVol~\cite{yeung2021implicitvol}
& Per-scene optimization
& 19.13 & 0.6128 & 0.5218
& \underline{18.97} & \underline{0.2837} & 0.7192 \\

UltraNeRF~\cite{wysocki2024ultra}
& Per-scene optimization
& 18.49 & 0.4056 & 0.5761
& 17.75 & 0.1975 & \textbf{0.5985} \\

3DGS~\cite{kerbl20233d}
& Per-scene optimization
& \underline{22.55} & \underline{0.6695} & \underline{0.3884}
& 18.04 & 0.2532 & 0.6702 \\

UltraGauss~\cite{eid2025ultragauss}
& Per-scene optimization
& 22.06 & 0.6220 & 0.4750
& 15.01 & 0.1838 & 0.7057 \\

\midrule

UltraVoxelGS w/o RAA
& Feed-forward
& 19.76 & 0.5770 & 0.5130
& 18.25 & 0.2690 & 0.7860 \\

\textbf{UltraVoxelGS (Ours)}
& Feed-forward
& \textbf{24.29} & \textbf{0.8329} & \textbf{0.2384}
& \textbf{20.87} & \textbf{0.3138} & \underline{0.6190} \\

\bottomrule
\end{tabular}}
\end{table}

\begin{figure}[t]
\centering
\includegraphics[width=\linewidth]{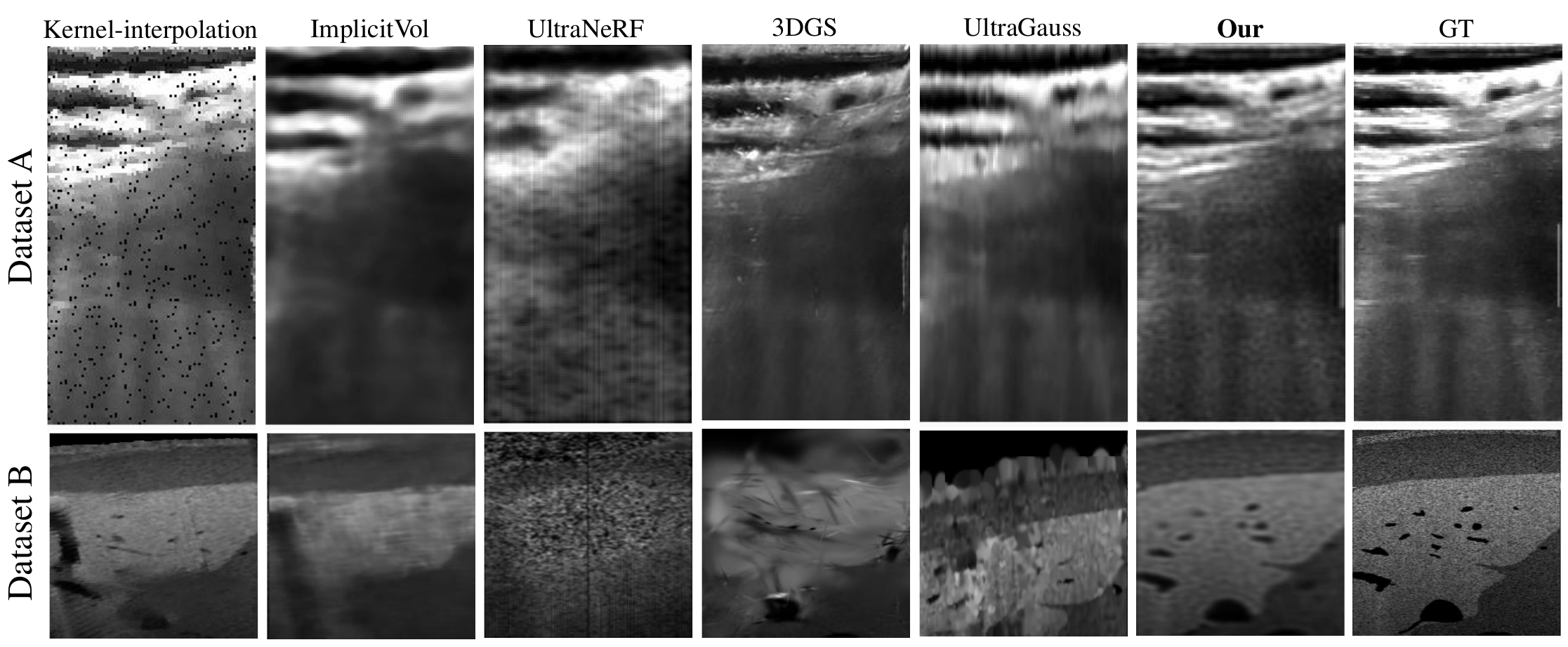}
\vspace{-0.2cm}
\caption{
{Qualitative comparison on real-world and simulated ultrasound datasets.}
}
\label{fig:qualitative_real}
\vspace{-0.5cm}
\end{figure}

\begin{figure}[ht!]
\centering
\includegraphics[width=\linewidth]{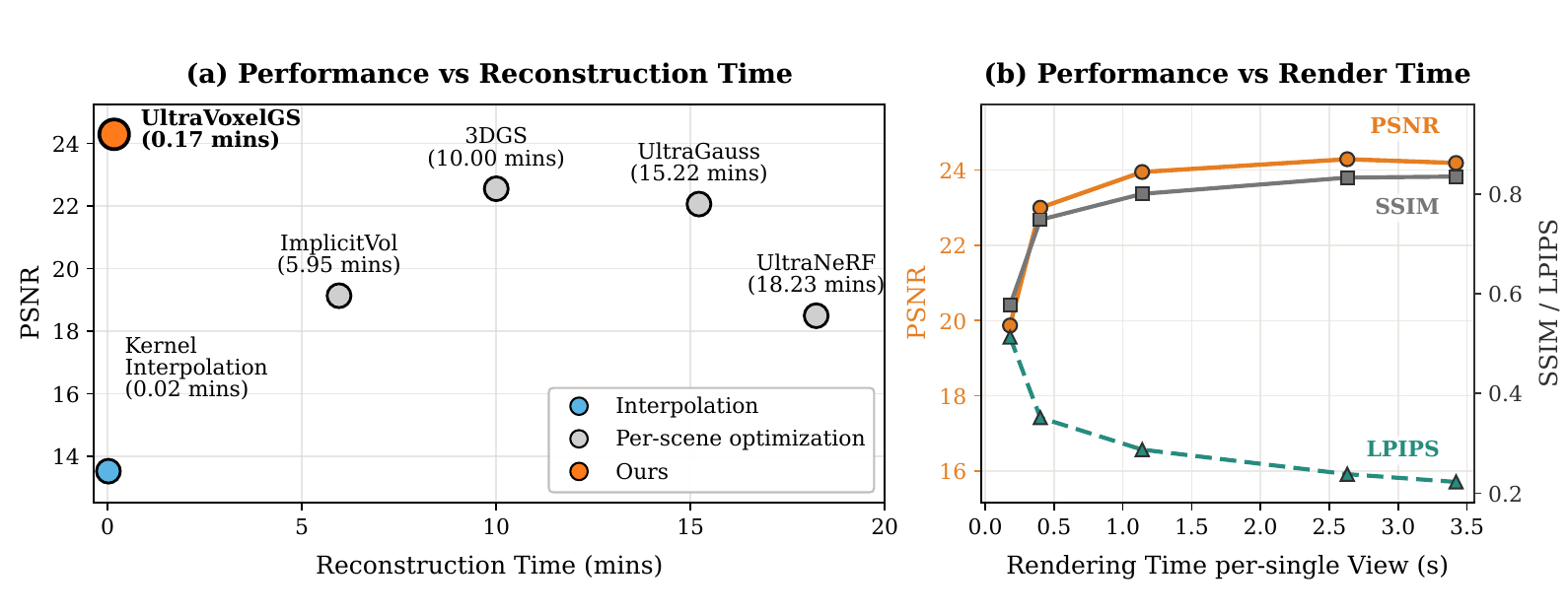}
\vspace{-0.8cm}
\caption{
{Quality--efficiency trade-off on Dataset A.}
}
\label{fig:quality_tradeoff}
\vspace{-0.3cm}
\end{figure}

\subsection{Main results on real-world and simulated data}


\textbf{Quantitative comparison.}
Tab.~\ref{tab:main_results} compares UltraVoxelGS with interpolation-based and per-scene optimization baselines on Datasets A and B.
UltraVoxelGS achieves the best overall performance on real tracked scans and remains competitive under controlled simulation, while using a feed-forward reconstruction pipeline \rev{for sub-second rendering} rather than per-scene optimization. These results support the effectiveness of the voxel-first formulation and ultrasound-aware appearance rendering for held-out slice reconstruction. \rev{The contribution of RAA is further analyzed in Appendix~\ref{disc: w/o RAA}, highlighting its role in compensating for fine details lost during feed-forward reconstruction.}


\textbf{Qualitative comparison.}  Fig.~\ref{fig:qualitative_real} shows representative qualitative comparisons on Dataset A and B. 
Interpolation-based reconstruction produces visible interpolation artifacts and loses anatomical detail. 
ImplicitVol yields overly smoothed responses, while UltraNeRF exhibits noisy and unstable texture patterns. 3DGS preserves some high-frequency appearance but introduces unstable speckle-like artifacts and local structural distortions. UltraGauss recovers part of the anatomical structure, but still shows degraded contrast and less coherent local organization. 
In comparison, UltraVoxelGS reconstructs clearer layered structures and more faithful local appearance.

\begin{figure}[t]
\centering
\includegraphics[width=\linewidth]{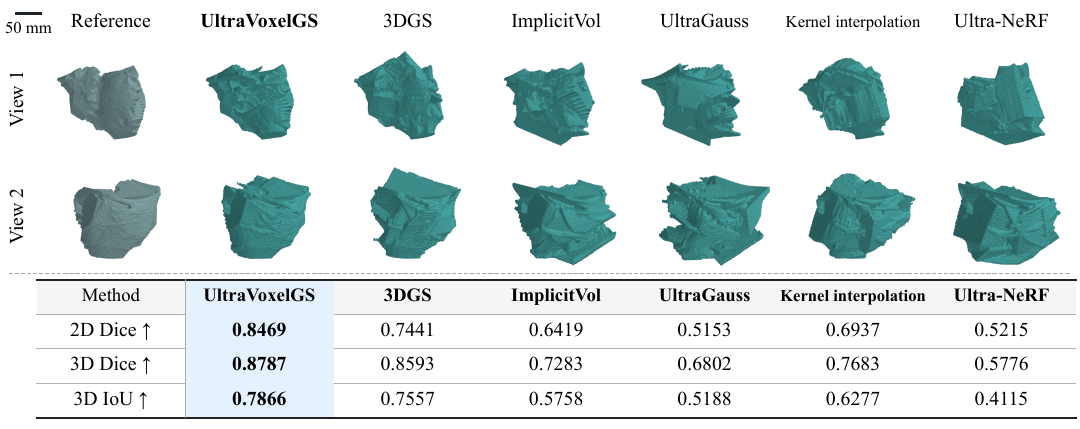 }
\caption{
3D anatomical segmentation results on Dataset B. Top: one representative example visualized from two viewpoints; Bottom: quantitative evaluation across all test cases.}
\vspace{-0.4cm}
\label{fig:3D_visual}
\end{figure}

\rev{\textbf{Segmentation-based anatomical consistency.} To assess 3D anatomical consistency, we evaluate on Dataset~B using CT-derived liver segmentation as the reference. For each method, we propagate an observed anchor-frame mask to the rendered novel views using the same frozen SAM2 model~\cite{ravi2025sam}, and fuse the resulting masks into a 3D anatomical reconstruction using the corresponding poses. As shown in Fig.~\ref{fig:3D_visual}, UltraVoxelGS achieves the highest 2D Dice, 3D Dice, and 3D IoU, with reconstructed anatomy most closely matching the reference shape, supporting its improved preservation of 3D anatomical consistency.}

\begin{figure}[t]
\centering
\includegraphics[width=\linewidth]{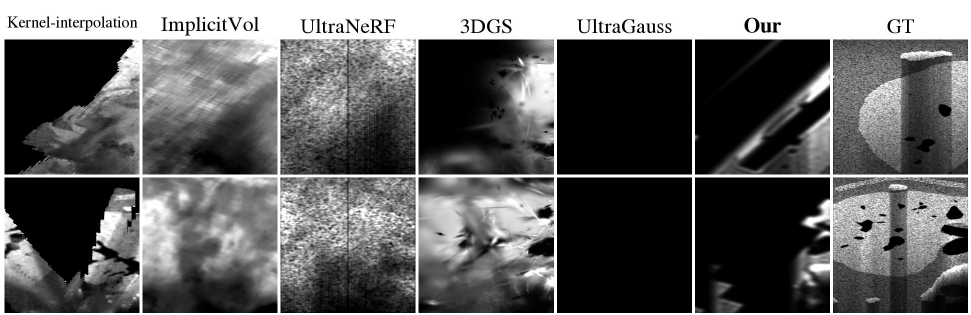}

\caption{\textbf{Failure cases on Dataset~B.} Visual comparisons under challenging scanning conditions, including sparse coverage, acoustic shadowing, and inaccurate probe poses.}
\vspace{-0.4cm}
\label{fig:failure_B_last}
\end{figure}

\textbf{Performance--efficiency trade-off.}
Fig.~\ref{fig:quality_tradeoff} (a) shows the relationship between quality and total per-scene reconstruction time on Dataset~A. 
UltraVoxelGS remains in the seconds-level reconstruction regime while matching or surpassing per-scene optimization baselines, showing that our method improves the full reconstruction pipeline rather than merely accelerating rendering. 
Fig.~\ref{fig:quality_tradeoff} (b) further analyzes the Render-time Appearance Adaptation trade-off: increasing adaptation steps improves PSNR and SSIM, but also increases per-view rendering cost. 
We use 50 adaptation steps in the main experiments to balance reconstruction quality and rendering efficiency. 
A detailed runtime breakdown is provided in Appendix~\ref{app:speed}.

\begin{table*}[t]
\centering
\begin{minipage}[t]{0.46\textwidth}
\centering
\caption{{Ablation study on Dataset A.}
We isolate the contributions of feed-forward (FF), Decoupled Reflection and Attenuation Rendering (DRA), and Render-time Appearance Adaptation (RAA).}
\label{tab:formulation}
\resizebox{\linewidth}{!}{
\begin{tabular}{ccc|ccc}
\toprule
FF & DRA & RAA
& PSNR$\uparrow$ & SSIM$\uparrow$ & LPIPS$\downarrow$ \\
\midrule

\xmark & \xmark & \xmark
& 13.52 & 0.3950 &0.5420\\

\cmark & \xmark & \xmark
& 14.46 & 0.3825 &{0.5735}\\

\cmark & \cmark & \xmark
& {19.76} & {0.5770} &0.5130\\

\cmark & \cmark & \cmark
& \textbf{24.29} & \textbf{0.8330} &\textbf{0.2380} \\ 
\bottomrule
\end{tabular}}
\end{minipage}
\hfill
\begin{minipage}[t]{0.5\textwidth}
\centering
\caption{
{Long-sequence scalability on Dataset A.}
UltraVoxelGS remains stable as input density increases from sparse to dense tracked scans, without retraining. Note that 'PM' indicates peak memory (GB) during inference.
}
\label{tab:long_sequence_scaling}
\resizebox{\linewidth}{!}{
\begin{tabular}{cccccc}
\toprule
Stride
& Frames
& PSNR$\uparrow$
& SSIM$\uparrow$
& LPIPS$\downarrow$
& PM$\downarrow$ \\

\midrule
10$\times$ & $\sim$300  
& 23.75 & 0.7940 & 0.2546 & \textbf{3.66} \\
5$\times$  
& $\sim$600  
& 24.28 & 0.8329 & 0.2386 & 3.80 \\
2$\times$  
& $\sim$1600 
& \textbf{24.89} & \textbf{0.8777} & \textbf{0.2175} & 3.94 \\
\bottomrule
\end{tabular}}
\end{minipage}
\end{table*}

\subsection{Ablation and analysis}



\textbf{Ablation study.}
Tab.~\ref{tab:formulation} isolates the contributions of feed-forward voxel prediction (FF), Decoupled Reflection and Attenuation Rendering (DRA), and Render-time Appearance Adaptation (RAA) on Dataset A. 
Starting from direct interpolation, introducing feed-forward voxel prediction improves PSNR but remains limited without a Gaussian scene representation and ultrasound-aware rendering. 
Adding DRA moves the model from voxel intensity prediction to slice-consistent Gaussian rendering with decoupled anisotropic reflection and cumulative attenuation, substantially improving PSNR, SSIM, and LPIPS. 
Finally, RAA provides the largest gain across all metrics by using nearby observed slices as appearance anchors for target-view refinement. 
Overall, the ablation supports the two main design choices of UltraVoxelGS: voxel-first feed-forward reconstruction and ultrasound-aware appearance rendering. \rev{We further compare UltraVoxelGS a generic feed-forward GS baseline, to assess the effectiveness of our ultrasound-specific design in Appendix~\ref{app:ablation_vol}.}

\textbf{Scalability analysis.}
We evaluate long-sequence scalability by varying the input sampling stride on Dataset A at reconstruction time, without retraining the model.
As shown in Tab.~\ref{tab:long_sequence_scaling}, UltraVoxelGS scales to over 1,000 input frames: denser inputs consistently improve reconstruction quality, while peak memory increases only mildly.
This supports the effectiveness of the proposed local-to-global assembly for dense tracked ultrasound sequences.

\begin{figure}[t]
\centering
\includegraphics[width=\linewidth]{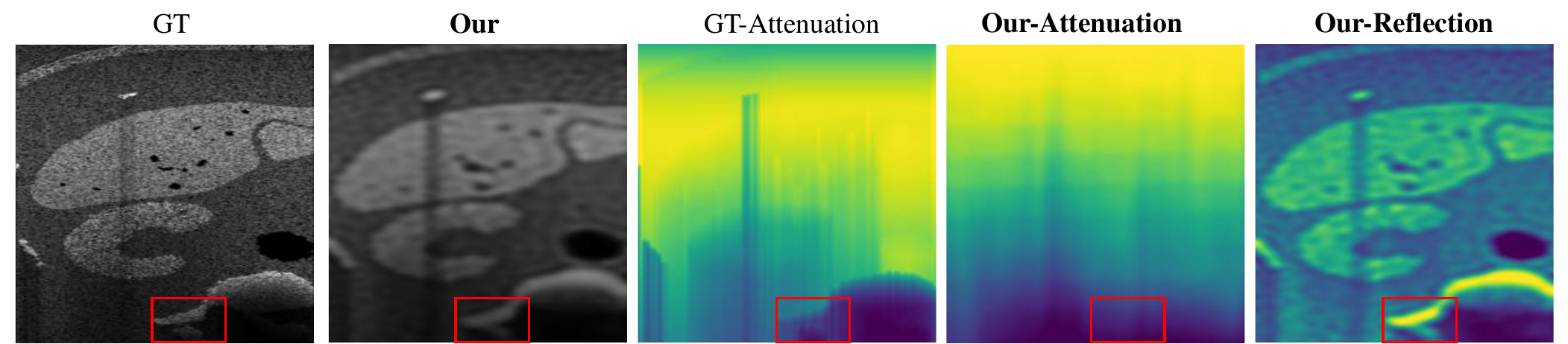}
\caption{
{Decoupled Reflection and Attenuation Rendering visualization on Dataset B.} Note that the ground-truth attenuation maps are obtained from the simulation pipeline \cite{velikova2023lotus} and serve solely as a visual reference; no attenuation supervision is used during training. The red boxes highlight a structure that is present in the reflection component but becomes weak in the final ultrasound appearance due to cumulative attenuation.
}
\label{fig:decoupled_attenuation}
\vspace{-0.4cm}
\end{figure}

\textbf{Attenuation and reflection representation.} 
Dataset~B provides simulator reference attenuation maps, which we use only as a visual reference rather than training supervision.
Fig.~\ref{fig:decoupled_attenuation} shows that UltraVoxelGS reconstructs the target slice while decomposing the rendered appearance into a cumulative attenuation component and an anisotropic reflection component. 
The learned attenuation component roughly follows the dominant depth-dependent trend in the simulator reference, while the reflection component captures complementary anatomical structures and local echo responses. 
\rev{Additionally, We quantitatively compare the learned attenuation component with the reference attenuation map: $\mathrm{Pearson}=0.6966\pm0.0259,\quad \mathrm{Spearman}=0.6699\pm0.0331$. Compared with a depth-linear decay baseline, whose slope and intercept are least-squares fitted to the reference ($\mathrm{Pearson}=0.6648$, $\mathrm{Spearman}=0.6545$), our results confirm that the factorization captures non-trivial attenuation structure despite receiving no supervision.}

\rev{\textbf{Accessibility of clinical workflow.} Fig.~\ref{fig:failure_B_last} shows representative failure cases under extremely sparse coverage, strong acoustic shadowing, or noticeable probe tracking errors. Compared with ImplicitVol and UltraNeRF, which may produce dense but over-smoothed or noisy responses in weakly observed regions, UltraVoxelGS tends to preserve evidence-supported structures while leaving under-covered regions visually incomplete. Such incompleteness can also provide immediate spatial feedback, indicating regions that may require additional scanning, as illustrated in Fig.~\ref{fig:runtime_workflow_pareto}. To further validate clinical accessibility, we additionally evaluate our method on a handheld dataset in Appendix~\ref{app:handheld}, further demonstrating its value in real-world settings. Moreover, the periodic reconstruction mode presented in Appendix~\ref{app:streaming} demonstrates its potential for continuous scanning.}


\section{Conclusion}


We have presented UltraVoxelGS, a feed-forward framework that, for the first time, achieves both fast reconstruction and physically faithful rendering for ultrasound imaging. The core insight is that when views are geometrically thin and photometrically inconsistent, the right strategy is to consolidate all observations into a shared volumetric representation first and predict locally. Paired with an ultrasound-aware appearance rendering strategy grounded in acoustic image formation, our design delivers reconstruction quality surpassing per-scene optimization baselines while reducing inference time from minutes to seconds and scaling to over 1,000 frames. Looking ahead, we hope UltraVoxelGS can enable real-time spatial feedback during ultrasound acquisition, guiding sonographers toward uncovered regions. Limitations are discussed in Appendix ~\ref{app:limitations}.

\rev{\section{Acknowledgement}
This work is supported by Hong Kong Research Grants Council (RGC) Early Career Scheme grant 22203525. The work is also supported by Guangdong and Hong Kong Universities “1+1+1” Joint Research Collaboration Scheme. The authors declare no competing interests.}

\bibliographystyle{ieeetr}
\bibliography{refs}
\medskip



\appendix

\section*{Appendix}
\addcontentsline{toc}{section}{Appendix}

\rev{We provide additional experimental details, runtime and RAA analyses, ablation and robustness studies, reconstruction results, clinical workflow discussions, and limitations to supplement the main paper.}

\section{Experimental Details}
\label{app:exp_details}




\subsection{Implementation details.}
UltraVoxelGS first constructs a unified voxel support from posed ultrasound slices and then predicts voxel-aligned Gaussian primitives from view-interacted local regions. 
Unless otherwise specified, the input image resolution is $176\times88$, the voxel grid spacing is $0.00075\,\mathrm{m}$, and each voxel patch has size $8\times8\times8$. 
For voxel support construction, each voxel uses up to $k=3$ nearest lifted observations. 
The Gaussian primitive is parameterized as $(\mathbf{c}_g,\Sigma_g,\boldsymbol{\rho}_g,\beta_g)$, where $\boldsymbol{\rho}_g\in\mathbb{R}^{9}$ corresponds to second-order spherical-harmonic coefficients for anisotropic reflection and $\beta_g$ denotes the attenuation coefficient.

We train UltraVoxelGS using Adam with learning rate $1\times10^{-5}$ for 300k iterations on one NVIDIA H20 GPU. 
The batch size is varied between 2 and 12 according to the number of selected view-interacted local regions and GPU memory. 
The overlap consistency weight is set to $\lambda=0.01$. 
Render-time Appearance Adaptation (RAA) is disabled during training and is applied only during target-view rendering.

\begin{table}[h]
\centering
\caption{\textbf{Implementation details of UltraVoxelGS.}}
\label{tab:implementation_details}
\resizebox{0.65\linewidth}{!}{
\begin{tabular}{lc}
\toprule
Item & Setting \\
\midrule
Input image resolution & $176\times88$ \\
Voxel grid spacing & $0.00075\,\mathrm{m}$ \\
Voxel patch size & $8\times8\times8$ \\
Top-$k$ neighbors & 3 \\
SH order & 2 \\
Gaussian parameters & $(\mathbf{c}_g,\Sigma_g,\boldsymbol{\rho}_g,\beta_g)$ \\
Optimizer & Adam \\
Learning rate & $1\times10^{-5}$ \\
Training steps & 300k \\
Batch size & 2--12 \\
Overlap consistency weight $\lambda$ & 0.01 \\
Training GPU & 1$\times$ NVIDIA H20 \\
\bottomrule
\end{tabular}}
\end{table}





\subsection{Baseline Settings}

We compare UltraVoxelGS with representative interpolation-based, continuous-field, neural rendering, and optimization-based Gaussian reconstruction methods. 
All methods use the same input and supervision frames and are evaluated on the same held-out target views whenever applicable. 
For optimization-based baselines, we follow their default reconstruction pipelines and report total per-scene reconstruction time under the same evaluation protocol.

\begin{table}[h]
\centering
\caption{\textbf{Baseline settings.}
We summarize the main baseline configurations used for comparison.}
\label{tab:baseline_settings}
\resizebox{\linewidth}{!}{
\begin{tabular}{lccc}
\toprule
Method & Type & Input Setting & Optimization / Inference Setting \\
\midrule
Kernel Interpolation & Interpolation & Same sparse input & No optimization \\
ImplicitVol & Per-scene optimization & Same sparse input & 50k steps \\
UltraNeRF & Per-scene optimization & Same sparse input & 10k steps \\
3DGS & Per-scene optimization & Same sparse input & 10k steps \\
UltraGauss & Per-scene optimization & Same sparse input & 10k steps \\
UltraVoxelGS w/o RAA & Feed-forward & Same sparse input & Feed-forward only \\
UltraVoxelGS & Feed-forward & Same sparse input & Feed-forward + RAA \\
\bottomrule
\end{tabular}}
\end{table}

\subsection{Additional Details for Long-sequence Evaluation}
\label{app:scaling}

For the long-sequence scalability experiment in the main paper, we vary the input sampling stride on Dataset A without retraining the model. 
The original robotic scans contain over 3K tracked frames. 
We use 10$\times$, 5$\times$, and 2$\times$ sampled inputs, corresponding to approximately 300, 600, and 1,600 input frames, respectively. 
All settings are evaluated on the same held-out subset with view index $3 \bmod 10$, containing 2,047 novel views from 8 scans. 
This protocol isolates reconstruction-time scalability under increasingly dense inputs while keeping the trained model fixed.

\section{Runtime and Render-time Adaptation Analysis}
\label{app:runtime}

\subsection{Runtime Breakdown}
\label{app:speed}

\begin{table}[h]
\centering
\caption{\textbf{Runtime breakdown of UltraVoxelGS.}
We report per-scene representation construction time and average per-view rendering time on Dataset A.}
\label{tab:runtime_breakdown}
\begin{tabular}{lc}
\toprule
Stage & Time (s) $\downarrow$ \\
\midrule
Voxel support construction & 1.12 \\
View-interacted block extraction & 1.96 \\
Network inference & 4.83 \\
Gaussian fusion & 2.13 \\
\midrule
Representation construction & 10.04 \\
Rendering per target view w/o RAA & 0.00362 \\
Rendering per target view w/ 50-step RAA & 0.623 \\
\bottomrule
\end{tabular}
\end{table}

UltraVoxelGS constructs the Gaussian representation in 10.04 s per case. 
The representation construction time includes voxel support construction, view-interacted block extraction, feed-forward network inference, and Gaussian fusion. 
Render-time Appearance Adaptation does not change the reconstructed global representation; it only increases the per-view rendering cost. 
\rev{The average rendering time is 3.62\,ms per target view without RAA (2.53\,ms rasterization + 1.09\,ms filtering) and 623\,ms per target view with 50-step RAA. Rendering time is averaged over 651 }held-out target views, sharded across three V100 GPUs for throughput measurement.

\subsection{Render-time Appearance Adaptation Step Ablation}
\label{app:RAA_steps}

\begin{figure}[t]
\centering
\includegraphics[width=\linewidth]{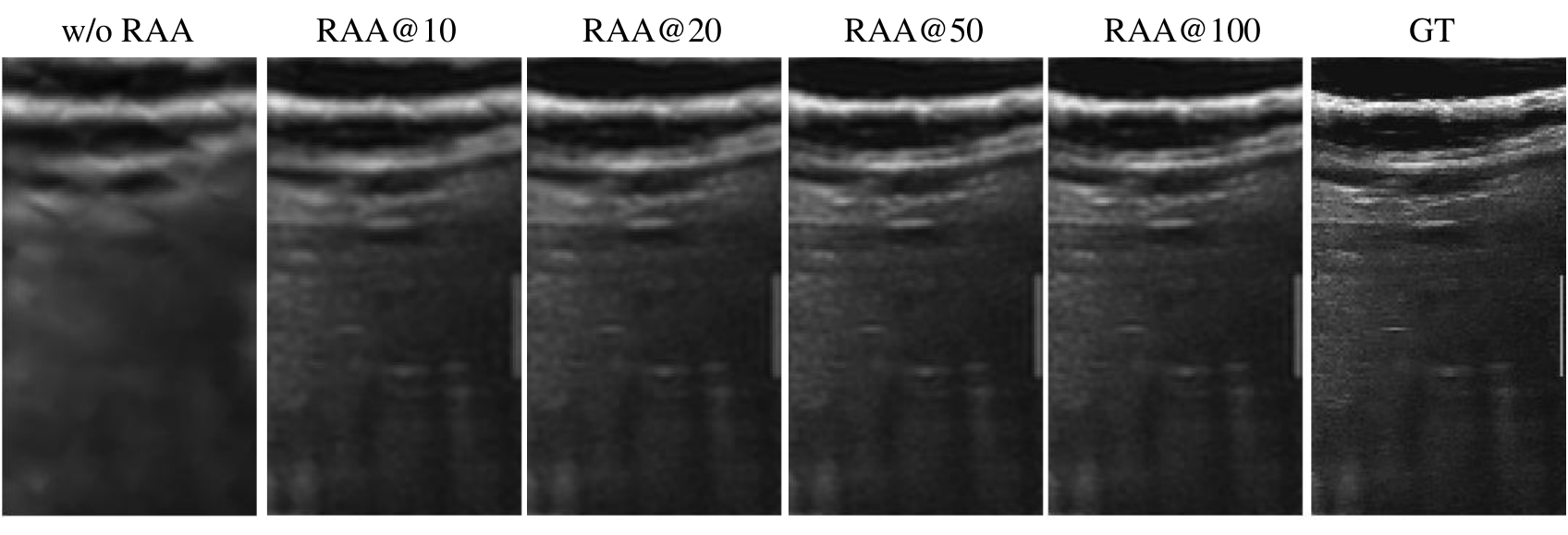}
\caption{
\textbf{Visual effect of Render-time Appearance Adaptation.}
RAA@$K$ denotes applying $K$ steps of Render-time Appearance Adaptation.
Increasing the number of RAA steps progressively improves local sharpness and contrast toward the ground truth.
}
\label{fig:raa_steps_visual}
\end{figure}

\textbf{Effect of adaptation steps.}
We evaluate the quality--efficiency trade-off of Render-time Appearance Adaptation (RAA) by varying the number of update steps.
For all variants, the feed-forward geometry is fixed, and only a transient copy of the appearance parameters is updated for each target rendering and discarded afterward.
As shown in Tab.~\ref{tab:RAA_steps}, more adaptation steps generally improve reconstruction quality at the cost of higher per-view rendering time.
Fig.~\ref{fig:raa_steps_visual} further shows the corresponding visual trend, where RAA progressively sharpens local structures and contrast toward the ground truth.


\begin{table}[h]
\centering
\caption{\textbf{Render-time Appearance Adaptation step ablation.}
We evaluate the quality--efficiency trade-off of different adaptation steps on Dataset A.}
\label{tab:RAA_steps}
\resizebox{0.75\linewidth}{!}{
\begin{tabular}{lccccc}
\toprule
Variant
& Steps
& PSNR$\uparrow$
& SSIM$\uparrow$
& LPIPS$\downarrow$
& Render / View (s)$\downarrow$ \\

\midrule
w/o RAA & 0  & 19.76 & 0.5770 & 0.5130 & \textbf{0.00362} \\
RAA & 10 & 23.00 & 0.7484 & 0.3513 & 0.140 \\
RAA & 20 & 23.95 & 0.8003 & 0.2872 & 0.259 \\
RAA & 50 & \textbf{24.29} & 0.8329 & 0.2380 & 0.623 \\
RAA & 100 & 24.19 & \textbf{0.8350} & \textbf{0.2224} & 1.212 \\
\bottomrule
\end{tabular}}
\end{table}


Increasing the number of adaptation steps improves local appearance fidelity but also increases per-view rendering time. 
We use 50 adaptation steps in the main experiments to balance reconstruction quality and rendering efficiency.

\subsection{RAA Sensitivity to Target–Anchor Discrepancy}

\rev{To analyze the sensitivity of RAA to pose variation, we rank all candidate anchors for each target view by either translational distance or rotational discrepancy, and apply RAA using anchors at different ranks (Top-1, Top-3, Top-5, and Top-10). As shown in Tab.~\ref{tab:anchor_ranking_ablation}, RAA maintains strong refinement within a small spatial and angular neighborhood, and although the benefit gradually decreases as the anchor's translational or rotational discrepancy grows, it consistently outperforms UltraVoxelGS w/o RAA even at Top-10.}


\begin{table}[t]
\centering
\small
\caption{RAA refinement under translation- and rotation-ranked anchors at different Top-$k$ ranks on Dataset~B.}
\label{tab:anchor_ranking_ablation}
\setlength{\tabcolsep}{4pt}
\resizebox{\linewidth}{!}{%
\begin{tabular}{llccccc}
\toprule
Strategy & Anchor
& Translation distance (mm)
& Rotation angle ($^\circ$)
& PSNR$\uparrow$
& SSIM$\uparrow$
& LPIPS$\downarrow$ \\
\midrule

UltraVoxelGS
& Top-1
& $7.87\pm3.52$
& $2.78\pm2.31$
& 20.87
& 0.3138
& 0.6190 \\

\midrule

\multirow{3}{*}{Translation}
& Top-3
& $14.09\pm5.56$
& $5.16\pm3.87$
& 19.89
& 0.2889
& 0.6919 \\

& Top-5
& $19.09\pm6.83$
& $6.80\pm4.47$
& 19.39
& 0.2813
& 0.7018 \\

& Top-10
& $28.86\pm8.27$
& $10.28\pm6.12$
& 18.76
& 0.2719
& 0.7136 \\

\midrule

\multirow{3}{*}{Rotation}
& Top-3
& $16.98\pm10.25$
& $4.11\pm1.95$
& 19.91
& 0.2889
& 0.6922 \\

& Top-5
& $22.24\pm10.46$
& $5.59\pm2.39$
& 19.41
& 0.2812
& 0.7020 \\

& Top-10
& $34.32\pm14.24$
& $8.45\pm3.10$
& 18.68
& 0.2707
& 0.7156 \\

\midrule

UltraVoxelGS w/o RAA
& -
& -
& -
& 18.25
& 0.2690
& 0.7860 \\

\bottomrule
\end{tabular}%
}
\end{table}


\subsection{Comparison with Per-scene Optimization}


\rev{RAA differs from short per-scene optimization in two aspects (shown in Tab.~\ref{tab:r2_raa}). First,  they differ in \emph{what is optimized}: RAA adapts transient appearance parameters while keeping the reconstructed geometry frozen, whereas per-scene optimization updates the full scene representation itself. Second, they differ in \emph{how the cost scales}: RAA is a query-time refinement with a fixed per-query cost that is independent of the number of source views, whereas per-scene optimization scales with the length of the input sequence, making it ill-suited to long ultrasound acquisitions. Overall, RAA is not a shortened per-scene reconstruction but a query-time appearance adaptation, offering a more practical refinement strategy for interactive ultrasound inspection.}

\begin{table}[t]
\centering
\small
\caption{Per-scene optimization versus RAA.}
\label{tab:r2_raa}
\setlength{\tabcolsep}{4pt}
\resizebox{\linewidth}{!}{%
\begin{tabular}{lcc}
\toprule
& Per-scene optimization & RAA \\
\midrule
Target & Complete scene & Current queried view \\
Updated variables & Geometry and appearance & Transient $\rho,\beta$ \\
Global representation & Optimized per scene & Feed-forward and fixed \\
Scope & Entire scene & Query-relevant local support \\
Persistence & Stored in the scene & Discarded after rendering \\
Runtime & Increases with scene scale and input views & Independent of source views (623 ms / 50 steps) \\
\bottomrule
\end{tabular}%
}
\end{table}

\subsection{Role of RAA Across Different Acquisition Densities}
\label{disc: w/o RAA}



\rev{The feed-forward pass and RAA are complementary: whenever interpolation smooths out detail, RAA restores it—decisively when the effect is severe (Dataset~A) and as a consistent refinement when it is mild (Dataset~B). RAA is thus an integral component, not an optional add-on, keeping UltraVoxelGS robust across acquisition densities.}

\rev{On Dataset~A, the dense, closely-spaced poses cause interpolation to oversmooth fine structures and degrade the feed-forward prediction; as the dense acquisition also places anchors close to the target, RAA effectively recovers the lost detail, yielding a large gain of about 4~dB (shown in Tab. \ref{tab:main_results}). On Dataset~B, the sparse poses induce far less smoothing and the feed-forward prediction is already sharp, matching the other baselines; RAA still compensates for the residual interpolation-induced detail loss, contributing a smaller but still substantial gain of about 2~dB.}

\section{Additional Ablation and Robustness Studies}
\label{app:ablations}

\subsection{Supporting Analysis of Decoupled Reflection and Attenuation Rendering}
\label{app:physics_support}

We compare UltraGauss with an optimized variant of our representation, UltraVoxelGS-opt, to examine the proposed rendering parameterization under per-scene optimization. 
UltraGauss uses a coupled ultrasound rendering model, while UltraVoxelGS-opt decouples reflection and cumulative attenuation and models anisotropic reflection with SH coefficients.
Since the two methods differ in representation and initialization, this comparison is intended as supporting evidence rather than an isolated causal ablation.

\begin{table}[h] 
\centering 
\caption{\textbf{Supporting analysis of Decoupled Reflection and Attenuation Rendering.}
We compare UltraGauss with an optimized variant of our representation under per-scene optimization.}
\label{tab:physics}
\resizebox{0.85\linewidth}{!}{
\begin{tabular}{lcccc}
\toprule
Rendering Variant
& Decoupled $(\rho,\beta)$
& SH Reflection
& PSNR$\uparrow$
& SSIM$\uparrow$ \\
\midrule
UltraGauss
& \xmark
& \xmark
& 22.06
& 0.6220 \\
\textbf{UltraVoxelGS-opt}
& \cmark
& \cmark
& \textbf{22.55}
& \textbf{0.6245} \\
\bottomrule
\end{tabular}}
\end{table}

Tab.~\ref{tab:physics} shows that UltraVoxelGS-opt achieves comparable or slightly better reconstruction quality than UltraGauss, suggesting that the decoupled and direction-aware appearance parameterization remains competitive under optimization.

\subsection{Overlap Consistency Ablation}
\label{app:consistency}

\begin{table}[h]
\centering
\caption{\textbf{Overlap consistency ablation.}
We evaluate whether overlap consistency stabilizes repeated local predictions under the proposed local-to-global assembly. RAA is disabled in this ablation.}
\label{tab:consistency}
\resizebox{0.82\linewidth}{!}{
\begin{tabular}{lcccc}
\toprule
Variant
& Consistency Loss
& RAA
& PSNR$\uparrow$
& SSIM$\uparrow$ \\
\midrule
Block-wise only
& \xmark
& \xmark
& 19.85 & 0.5742 \\
Block-wise + Consistency
& \cmark
& \xmark
& \textbf{19.76} & \textbf{0.5770} \\
\bottomrule
\end{tabular}}
\end{table}

\textbf{Effect of overlap consistency.}
In local-to-global reconstruction, the same spatial support can be selected by multiple supervised views and thus receive repeated Gaussian predictions. 
We introduce overlap consistency to regularize these duplicated predictions before fusion. 
Tab.~\ref{tab:consistency} compares block-wise UltraVoxelGS with and without this loss, both without Render-time Appearance Adaptation. 
With overlap consistency, both PSNR and SSIM improve slightly, indicating that regularizing repeated local predictions makes the fused Gaussian representation more stable without degrading reconstruction fidelity. 
This supports its role in stabilizing global fusion across overlapping local predictions.


\subsection{Voxel Support Resolution Analysis}


\rev{To examine whether a sharper spatial support can reduce the effect of RAA, we increase the voxel resolution at inference time without changing the model or retraining it. As shown in Tab.~\ref{tab:r2_support}, the denser support provides a modest improvement for UltraVoxelGS w/o RAA, but increases the Gaussian count from approximately 38.0M to 90.5M. Meanwhile, the complete UltraVoxelGS (w/ RAA) still achieves substantially better performance.}


\begin{table}[h]
\centering
\small
\caption{ Expressive support-density analysis.}
\label{tab:r2_support}
\begin{tabular}{llccc}
\toprule
Support resolution & Method
& PSNR$\uparrow$ & SSIM$\uparrow$ & LPIPS$\downarrow$ \\
\midrule
$0.75$ mm & UltraVoxelGS w/o RAA
& 19.76 & 0.5770 & 0.5130 \\

$0.75$ mm & UltraVoxelGS 
& \textbf{24.29} & \textbf{0.8329} & \textbf{0.2384} \\
$0.50$ mm & UltraVoxelGS w/o RAA
& 20.14 & 0.5909 & 0.4719 \\

$0.50$ mm & UltraVoxelGS 
& 24.22 & 0.8200 & \textbf{0.2131} \\
\bottomrule
\end{tabular}
\end{table}

\subsection{Comparison with Generic Feed-Forward GS}
\label{app:ablation_vol}

\rev{To evaluate the effectiveness of our ultrasound-specific feed-forward reconstruction paradigm, we introduce VolSplat\cite{wang2025volsplat}, a generic feed-forward GS baseline, for comparison. However, directly applying VolSplat to long ultrasound sequences is computationally impractical because its native formulation jointly processes a limited number of input views before lifting their features into a 3D feature space and decoding the scene.}

\rev{We therefore adapt VolSplat to our long-sequence setting while preserving its original training paradigm. During training, randomly sampled source views are used as input, their features are lifted into a 3D feature space, and the complete scene is decoded. During inference, we divide all source views into groups, process each group independently to obtain lifted 3D features, fuse the features from all groups into a common global feature space, and decode the scene representation once. This allows VolSplat to use the full source sequence.}

\rev{The results on Dataset~A are shown in Tab.~\ref{tab:r2_volsplat}. UltraVoxelGS w/o RAA substantially outperforms the adapted VolSplat baseline. This gap reflects the distinct reconstruction paradigms. VolSplat must process limited view subsets independently, losing the global scene context and making it difficult to aggregate information distributed across different subset views. In contrast, UltraVoxelGS first consolidates the complete sequence into a unified 3D support and then predicts Gaussians only from the view-interacted local regions. Owing to ultrasound thin-slab locality, each such local region contains the relevant context from the globally constructed support without requiring repeated whole-scene decoding. The result therefore supports the contribution of the ultrasound-specific voxel-first formulation and local-to-global reconstruction strategy.}

\begin{table}[h]
\centering
\small
\caption{Feed-forward comparison on Dataset~A.}
\label{tab:r2_volsplat}
\begin{tabular}{lcccc}
\toprule
Method & Regime & PSNR$\uparrow$ & SSIM$\uparrow$ & LPIPS$\downarrow$ \\
\midrule
VolSplat
& Feed-forward
& 15.70 & 0.4522 & 0.5884 \\

UltraVoxelGS w/o RAA
& FF + DRA
& 19.76 & 0.5770 & 0.5130\\

UltraVoxelGS
& FF + DRA + RAA
& 24.29 & 0.8329 & 0.2384 \\
\bottomrule
\end{tabular}
\end{table}

\subsection{Controlled pose-noise Analysis}

\rev{The purpose of this experiment is to evaluate whether UltraVoxelGS remains stable when the input poses contain tracking errors. Since training data can be curated with offline pose refinement, we focus on the clinically relevant scenario of inference-time tracking noise. Pose information is used in three stages of our framework: constructing the global representation (fusion), selecting the observed anchor (anchor), and querying the target view (query).}

\rev{To isolate the sensitivity of each stage, we decompose the effect across four conditions: (A) clean setting; (B) noisy poses are used only for scene fusion; (C) noisy poses are used for both scene fusion and anchor selection; and (D) noisy poses are used for all three stages. Specifically, we inject zero-mean Gaussian noise independently for each frame and for each translational and rotational axis at two levels ($\sigma_t=8$\,mm/$\sigma_r=4^\circ$ and $\sigma_t=12$\,mm/$\sigma_r=6^\circ$), and report the mean and standard deviation over three independent random seeds. As shown in Tab.~\ref{tab:r1_pose_noise}, fusion is essentially unaffected (B remains at the clean-pose level, within run-to-run variation), and the degradation is concentrated in anchor selection, while target querying adds little. Notably, even under such large perturbations (realized RMS $20.8$\,mm/$10.3^\circ$), PSNR drops by only $\sim$1.38\,dB, demonstrating that UltraVoxelGS is robust to realistic tracking errors.}

\begin{table*}[h]
\centering
\scriptsize
\setlength{\tabcolsep}{3.2pt}
\caption{Reproducibility of controlled pose experiment on dataset~B.}
\label{tab:r1_pose_noise}
\begin{tabular}{cc l ccc}
\toprule
$\sigma_t$ (mm) & $\sigma_r$ ($^\circ$) & Condition
& PSNR$\uparrow$ & SSIM$\uparrow$ & LPIPS$\downarrow$ \\
\midrule
0 & 0 & A
& $20.86$
& $0.3158$
& $0.6206$ \\
\midrule
8 & 4 & B
& $20.92 \pm 0.01$
& $0.3177 \pm 0.0001$
& $0.6199 \pm 0.0001$ \\
& & C
& $20.11 \pm 0.08$
& $0.3017 \pm 0.0015$
& $0.6336 \pm 0.0013$ \\
& & D
& $19.92 \pm 0.08$
& $0.2985 \pm 0.0016$
& $0.6384 \pm 0.0043$ \\
\midrule
12 & 6 & B
& $20.87 \pm 0.10$
& $0.3165 \pm 0.0027$
& $0.6232 \pm 0.0052$ \\
& & C
& $19.62 \pm 0.01$
& $0.2920 \pm 0.0007$
& $0.6423 \pm 0.0004$ \\
& & D
& $19.48 \pm 0.03$
& $0.2900 \pm 0.0002$
& $0.6448 \pm 0.0016$ \\
\bottomrule
\end{tabular}
\end{table*}



\section{Additional Reconstruction Results}
\label{app:qualitative_failure}

\subsection{Additional Qualitative Results}

\begin{figure}[t]
\centering
\includegraphics[width=\linewidth]{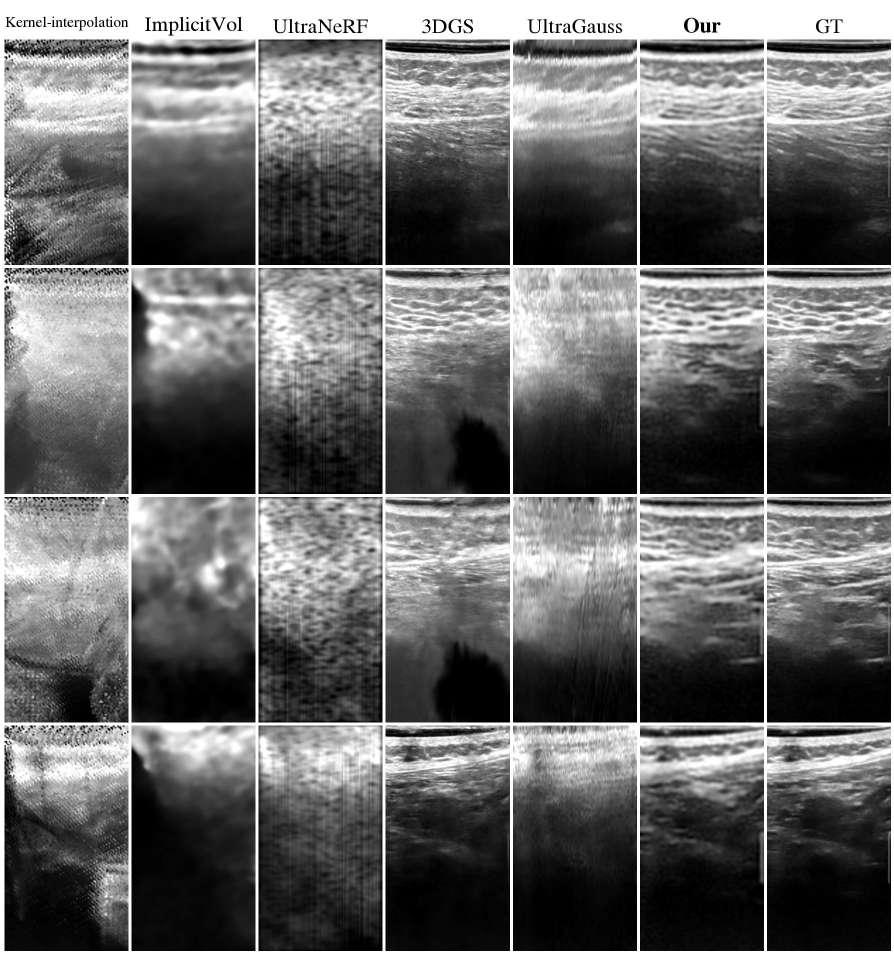}
\caption{
\textbf{Additional qualitative comparison on dataset~A.}
UltraVoxelGS better preserves anatomical structures, local sharpness, and cross-view consistency under sparse ultrasound observations.
}
\label{fig:qualitative_A}
\end{figure}

\begin{figure}[t]
\centering
\includegraphics[width=\linewidth]{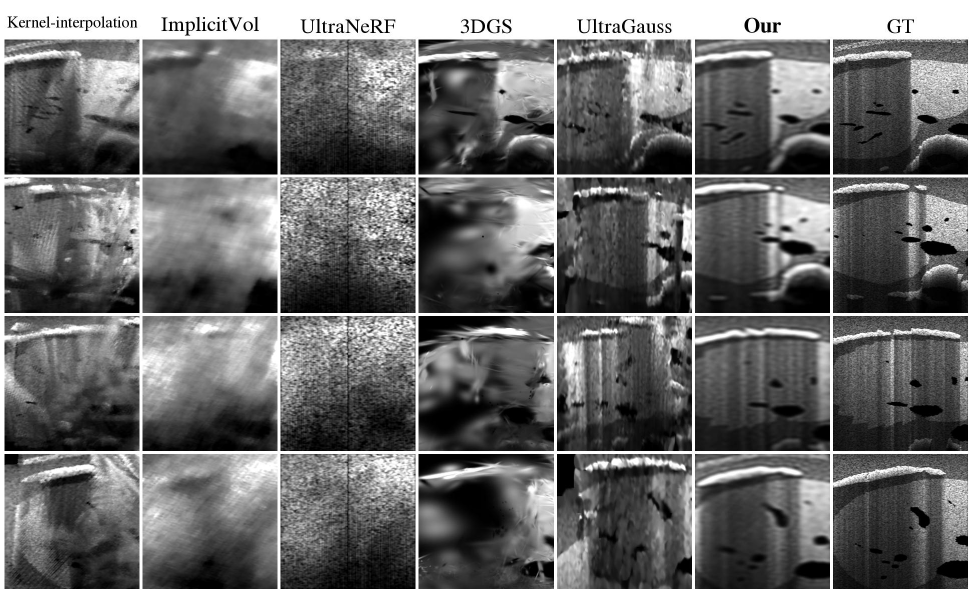}
\caption{
\textbf{Additional qualitative comparison on datasetB.}
UltraVoxelGS better preserves anatomical structures, local sharpness, and cross-view consistency under sparse ultrasound observations.
}
\label{fig:qualitative_B}
\end{figure}

Fig.~\ref{fig:qualitative_A} and Fig.~\ref{fig:qualitative_B} provides additional qualitative comparisons on held-out slice reconstruction. 
Compared with interpolation-based and optimization-based baselines, UltraVoxelGS better preserves layered anatomical structures and local appearance details under sparse ultrasound observations.

\subsection{Validation on Handheld Ultrasound}
\label{app:handheld}



\rev{To directly evaluate the handheld setting in real-world, we test the trained model on tracked-handheld Dataset~C---paired with Dataset~A but differing in acquisition trajectory---without retraining and under the same protocol as Dataset~A (Tab.~\ref{tab:r3_handheld}). No baseline is optimized for this handheld setting, yet UltraVoxelGS remains close to its own robotic performance and still outperforms the per-scene UltraGauss, confirming that it generalizes to the more complex trajectories of real-world handheld scanning.}

\begin{table}[h]
\centering
\small
\caption{Subject-level average on robotic and tracked-handheld test cases.}
\label{tab:r3_handheld}
\begin{tabular}{lccc}
\toprule
Setting & PSNR$\uparrow$ & SSIM$\uparrow$ & LPIPS$\downarrow$ \\
\midrule
UltraGauss (Handheld acquisition)
& 20.05 & 0.5858 & 0.5580 \\

UltraVoxelGS (Robotic acquisition)
& 24.29 & 0.8329 & 0.2384 \\

UltraVoxelGS (Handheld acquisition)
& 23.96 & 0.7703 & 0.3356 \\
\bottomrule
\end{tabular}
\end{table}


\section{Streaming and Clinical Workflow}

\rev{\subsection{Streaming / periodic reconstruction}}
\label{app:streaming}

As a natural extension of UltraVoxelGS, we have implemented and evaluated Streaming/periodic reconstruction. For each voxel~$\mathbf{x}$ we maintain two persistent fusion accumulators, the weighted numerator $A(\mathbf{x})$ and denominator $W(\mathbf{x})$:
\begin{equation}
A_t(\mathbf{x}) = \sum \omega_{i,p}(\mathbf{x})\,I_i(p), \qquad W_t(\mathbf{x}) = \sum \omega_{i,p}(\mathbf{x}),
\end{equation}
where the summation aggregates the contributions of all frames processed so far. When a new chunk of frames arrives, they add their contributions $(\Delta A_t, \Delta W_t)$ only to the voxels they affect, and the voxel intensity is the updated ratio:
\begin{equation}
A_{t+1} = A_t + \Delta A_t, \qquad W_{t+1} = W_t + \Delta W_t, \qquad V_{t+1}(\mathbf{x}) = \frac{A_{t+1}(\mathbf{x})}{W_{t+1}(\mathbf{x}) + \epsilon}.
\end{equation}
This incremental fusion follows directly from Eq.~\ref{eq:voxel_interpolation}. For the Gaussian stage, only the blocks whose voxel support has been updated by the current chunk, $\mathcal{U}_t = \{\mathbf{x} \mid \Delta W_t(\mathbf{x}) > 0\}$, are re-processed by the voxel-to-Gaussian network, since these are the only blocks that have received new information; all remaining blocks keep their previous Gaussians unchanged:
\begin{equation}
\mathbf{G}_{t+1}^{\,b} = \begin{cases} f_{\mathrm{V2G}}\!\left(V_{t+1}^{\,b}\right), & b \in \mathcal{B}_t,\\ \mathbf{G}_t^{\,b}, & b \notin \mathcal{B}_t. \end{cases}
\end{equation}

We evaluate the streaming mode in two ways. Compared with direct reconstruction, which builds the voxel support from the full sequence, generates the Gaussians block-wise, and evaluates all novel views at once, the two streaming settings proceed incrementally. In \emph{progressive streaming}, every $L$ incoming frames update the current voxel support (following the equations above), and only the blocks whose support has changed are re-processed by the feed-forward model to refresh the corresponding Gaussians. Right after each update, we evaluate the novel views whose target planes fall within the spatial extent of the newly updated blocks; each novel view is evaluated exactly once, immediately after the update that covers it, using only the information available at that point.

\emph{Final streaming} shares this exact construction but evaluates all novel views once at the end, using the representation obtained after the full sequence is processed. 

\begin{figure}[t]
\centering
\includegraphics[width=0.8\linewidth]{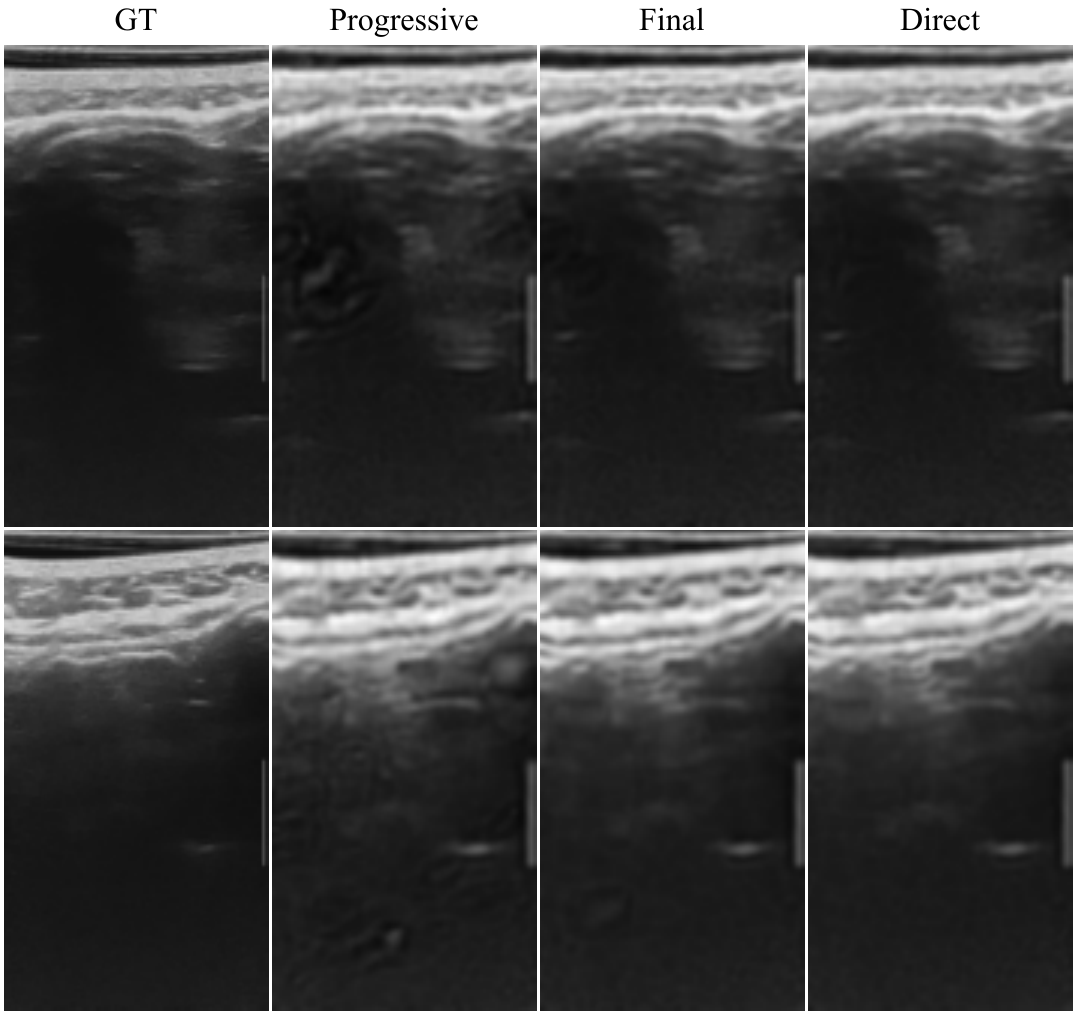}
\caption{\textbf{Qualitative comparison of reconstruction modes on Dataset~B.} Progressive and final streaming achieve visual quality comparable to direct reconstruction.}
\label{fig:progressive}
\end{figure}

In streaming experiments we set the chunk size to $L=10$ frames. As shown in Tab.~\ref{tab:streaming_reconstruction}, both settings are comparable to direct reconstruction, confirming that the online paradigm is practical without sacrificing fidelity. Fig.~\ref{fig:progressive} further shows that both streaming modes preserve comparable anatomical structures and visual details. Final streaming is slightly better than progressive streaming, as it evaluates every view using the complete sequence, whereas progressive streaming can only use the frames seen up to each point. Direct reconstruction performs slightly better than final streaming because the feed-forward model is trained under the direct (full-volume) setting; the streaming pipeline \textbf{reuses the same model without any retraining or fine-tuning}, so the minor gap is expected and reflects only the mismatch between training and inference procedures rather than a fundamental limitation of the streaming paradigm.

\begin{table}[h]
\centering
\small
\setlength{\tabcolsep}{5pt}
\caption{Performance of streaming reconstruction on dataset~A.}
\label{tab:streaming_reconstruction}
\begin{tabular}{lccc}
\toprule
Setting & PSNR$\uparrow$ & SSIM$\uparrow$ & LPIPS$\downarrow$ \\
\midrule
Progressive streaming & 24.08 & 0.8240 & 0.2555 \\
Final streaming & 24.11 & 0.8280 & 0.2476 \\
Direct reconstruction & \textbf{24.29} & \textbf{0.8329} & \textbf{0.2384} \\
\bottomrule
\end{tabular}
\end{table}

\rev{\subsection{Comparison with Commercial Reconstruction Systems}}

The key difference lies in the underlying scene representation. UltraVoxelGS uses a direction-aware Gaussian representation, which allows it to model ultrasound-specific appearance factors such as anisotropic reflection and cumulative attenuation. This representation better accommodates flexible scanning trajectories, including local revisits and changes in scanning direction.

In contrast, many conventional or commercial reconstruction systems form an isotropic voxel volume by lifting registered 2D slices into 3D space and compounding their intensities, producing a tomographic-style representation similar to CT or MRI volumes. Such reconstruction commonly favors smooth, approximately uniform sweeps with limited dwelling or repeated sampling of the same region, because redundant observations are often averaged during voxel compounding, which may smooth local details. This can constrain the acquisition trajectory and, consequently, the applicable scanning scenarios.

This difference is particularly relevant near interfaces with large acoustic-impedance differences, where strong specular reflection or acoustic shadowing may hide structures from a single insonification direction. Additional observations from different probe directions can therefore be important. UltraVoxelGS is designed to preserve and render such direction-dependent observations within a unified representation rather than collapsing them into a single isotropic intensity value.

\rev{\subsection{Visual Feedback and Scan Guidance}}

Real-time scan guidance requires two capabilities: (i) fast 3D reconstruction and rendering that keeps pace with the ongoing scan, and (ii) a mechanism to identify under-reconstructed regions and translate them into actionable probe-pose suggestions. This work addresses capability (i). UltraVoxelGS enables a live reconstruction workflow: as the operator scans, the system simultaneously streams acquired frames into a voxel support and, at fixed intervals, generates local GS representations and fuses them into the global scene. This construction proceeds in the background, overlapping with the scan rather than waiting for it. When inspection is needed, the operator can render corresponding views in real time.

Regarding capability (ii), reconstruction incompleteness is already visually observable in the rendered views (see Fig.~\ref{fig:qualitative_A}), but it has not yet been converted into explicit guidance. One natural extension is to associate uncertainty with the Gaussian representation and compute a pose-dependent cumulative uncertainty score for candidate poses, directing the operator toward under-observed regions. Beyond visual inspection, the reconstructed scene also supports downstream analysis such as 3D segmentation , which can further assist the operator in assessing anatomical coverage. These extensions are enabled by, and fundamentally depend on, the fast reconstruction and rendering prerequisite addressed in this work.

\section{Limitations and Broader Impact}
\label{app:limitations}

\textbf{Limitations.}
UltraVoxelGS relies on sufficiently accurate probe tracking to lift posed 2D ultrasound slices into a unified voxel support. 
The method may degrade when the observed coverage is extremely sparse, when severe acoustic shadowing removes local structural evidence, or when probe poses contain noticeable errors. 
In addition, our evaluation focuses on held-out slice reconstruction and representation quality, which does not directly establish clinical diagnostic performance. 
\rev{Bridging this gap would likely require explicit uncertainty estimation to flag unreliable regions and probe-pose guidance to direct operators toward uncovered anatomy, both of which we leave to future work.}

UltraVoxelGS also adopts a simplified ultrasound-aware appearance rendering model. 
Specifically, Decoupled Reflection and Attenuation Rendering focuses on anisotropic reflection and cumulative attenuation, which capture the main direction- and path-dependent factors in B-mode ultrasound intensity formation: local reflection determines the generated echo response, while cumulative attenuation controls signal decay along the acoustic path. 
This formulation is compatible with Gaussian scene representation, where Gaussian appearance parameters model local reflection/scattering and accumulated transmittance models attenuation. 
Our results on real-world and simulated datasets suggest that this approximation is effective for the reconstruction tasks studied in this work, but it is not a complete acoustic simulation. 
Other effects, such as speed-of-sound variation, refraction, speckle decorrelation, beamforming artifacts, and device-specific post-processing, are not explicitly modeled and may become important under large view changes, strong specular interfaces, severe shadowing, or cross-device settings. 
Future work may incorporate richer acoustic effects while preserving the efficiency of the feed-forward voxel-to-Gaussian pass.

\textbf{Broader impact.}
Fast 3D ultrasound reconstruction can support low-latency scan guidance, spatial feedback, and more efficient review of ultrasound acquisitions. By reducing reconstruction time and hardware requirements, feed-forward 3D ultrasound reconstruction may also help extend volumetric imaging capabilities to resource-limited clinical settings where dedicated 3D probes are unavailable. 
However, the current method should be viewed as a reconstruction and visualization tool rather than a standalone clinical decision system. 
Future work should further study uncertainty estimation, pose-noise robustness, broader freehand ultrasound validation, and prospective clinical evaluation.

\newpage
\input{checklist.tex}

\end{document}

%% file: checklist.tex
\section*{NeurIPS Paper Checklist}

\begin{enumerate}

\item {\bf Claims}
    \item[] Question: Do the main claims made in the abstract and introduction accurately reflect the paper's contributions and scope?
    \item[] Answer: \answerYes{} 
    \item[] Justification:  The abstract and introduction state the main contributions, including the voxel-first feed-forward Gaussian formulation, ultrasound-aware appearance rendering, Render-time Appearance Adaptation, and scalability to long ultrasound sequences. The claims are supported by the quantitative comparisons, ablation studies, runtime analysis, and the stated limitations. 

\item {\bf Limitations}
    \item[] Question: Does the paper discuss the limitations of the work performed by the authors?
    \item[] Answer: \answerYes{} 
    \item[] Justification: The paper includes a dedicated limitations discussion in the appendix. It discusses the dependence on accurate probe tracking, degradation under sparse coverage or severe acoustic shadowing, the simplified acoustic rendering model, and the fact that held-out slice reconstruction does not directly establish clinical diagnostic performance. 

\item {\bf Theory assumptions and proofs}
    \item[] Question: For each theoretical result, does the paper provide the full set of assumptions and a complete (and correct) proof?
    \item[] Answer: \answerYes{} 
    \item[] Justification: The paper does not present formal theoretical results, theorems, or proofs. The equations in the methodology define the reconstruction formulation, rendering model, and optimization objectives.

    \item {\bf Experimental result reproducibility}
    \item[] Question: Does the paper fully disclose all the information needed to reproduce the main experimental results of the paper to the extent that it affects the main claims and/or conclusions of the paper (regardless of whether the code and data are provided or not)?
    \item[] Answer: \answerYes{} 
    \item[] Justification:The paper describes the model formulation, rendering objective, training loss, datasets, evaluation splits, baseline settings, implementation details, and runtime protocol. Additional experimental details are provided in the appendix to support reproducibility of the reported comparisons. 

\item {\bf Open access to data and code}
    \item[] Question: Does the paper provide open access to the data and code, with sufficient instructions to faithfully reproduce the main experimental results, as described in supplemental material?
    \item[] Answer: \answerNo{} 
    \item[] Justification: Code and data will be released upon the acceptance of the paper.

\item {\bf Experimental setting/details}
    \item[] Question: Does the paper specify all the training and test details (e.g., data splits, hyperparameters, how they were chosen, type of optimizer) necessary to understand the results?
    \item[] Answer: \answerYes{} 
    \item[] Justification: 
The main paper specifies the datasets and held-out evaluation protocols, and the appendix provides implementation details including image resolution, voxel spacing, patch size, top-$k$ interpolation, SH order, optimizer, learning rate, training iterations, batch size, overlap consistency weight, and GPU setting.
    

\item {\bf Experiment statistical significance}
    \item[] Question: Does the paper report error bars suitably and correctly defined or other appropriate information about the statistical significance of the experiments?
    \item[] Answer: \answerNo{} 
    \item[] Justification: 
    The paper reports average quantitative metrics over held-out target views and cases, but does not currently include error bars, confidence intervals, or statistical significance tests. The evaluation uses consistent splits and case-level aggregation, but uncertainty estimates are not reported.

\item {\bf Experiments compute resources}
    \item[] Question: For each experiment, does the paper provide sufficient information on the computer resources (type of compute workers, memory, time of execution) needed to reproduce the experiments?
    \item[] Answer: \answerYes{} 
    \item[] Justification: The appendix reports the training GPU, training configuration, baseline optimization steps, runtime breakdown, per-view rendering cost with and without RAA, and peak memory for long-sequence scalability. These details describe the main compute resources required for reproducing the reported experiments.
    
\item {\bf Code of ethics}
    \item[] Question: Does the research conducted in the paper conform, in every respect, with the NeurIPS Code of Ethics \url{https://neurips.cc/public/EthicsGuidelines}?
    \item[] Answer: \answerYes{} 
    \item[] Justification: The work uses previously collected/cited ultrasound datasets and simulated ultrasound data, does not introduce new human-subject data collection, and does not deploy the method as a clinical decision system. The submission is anonymized and the paper discusses limitations and broader impact.

\item {\bf Broader impacts}
    \item[] Question: Does the paper discuss both potential positive societal impacts and negative societal impacts of the work performed?
    \item[] Answer: \answerYes{} 
    \item[] Justification: The appendix discusses the potential positive impact of fast 3D ultrasound reconstruction for low-latency scan guidance, spatial feedback, and resource-limited settings. It also notes that the method should not be treated as a standalone clinical diagnostic system and requires further robustness, uncertainty, and clinical validation.
    
\item {\bf Safeguards}
    \item[] Question: Does the paper describe safeguards that have been put in place for responsible release of data or models that have a high risk for misuse (e.g., pre-trained language models, image generators, or scraped datasets)?
    \item[] Answer: \answerNA{} 
    \item[] Justification: The paper poses no such risks

\item {\bf Licenses for existing assets}
    \item[] Question: Are the creators or original owners of assets (e.g., code, data, models), used in the paper, properly credited and are the license and terms of use explicitly mentioned and properly respected?
    \item[] Answer: \answerYes{} 
    \item[] Justification: The paper credits the original datasets, simulation tools, and baseline methods through citations, but the current manuscript does not exhaustively list the license names and terms of use for every existing asset. These details will be added to the release documentation and should be included in the final version where applicable.

\item {\bf New assets}
    \item[] Question: Are new assets introduced in the paper well documented and is the documentation provided alongside the assets?
    \item[] Answer: \answerNA{} 
    \item[] Justification: The submission does not introduce or release a new dataset at submission time. If code, trained models, or processed data splits are released after acceptance, they will be accompanied by documentation, usage instructions, licenses, and limitations.

\item {\bf Crowdsourcing and research with human subjects}
    \item[] Question: For crowdsourcing experiments and research with human subjects, does the paper include the full text of instructions given to participants and screenshots, if applicable, as well as details about compensation (if any)? 
    \item[] Answer: \answerNA{}
    \item[] Justification: The paper does not conduct crowdsourcing experiments or new human-subject studies. It uses previously collected/cited ultrasound data and simulated data, so no participant instructions, screenshots, or compensation details are applicable.

\item {\bf Institutional review board (IRB) approvals or equivalent for research with human subjects}
    \item[] Question: Does the paper describe potential risks incurred by study participants, whether such risks were disclosed to the subjects, and whether Institutional Review Board (IRB) approvals (or an equivalent approval/review based on the requirements of your country or institution) were obtained?
    \item[] Answer: \answerNA{} 
    \item[] Justification: The authors do not collect new human-subject data in this work. The real ultrasound data are previously collected/cited data, and the simulated dataset is generated from existing resources; therefore, new IRB approval for participant recruitment is not applicable to this submission.

\item {\bf Declaration of LLM usage}
    \item[] Question: Does the paper describe the usage of LLMs if it is an important, original, or non-standard component of the core methods in this research? Note that if the LLM is used only for writing, editing, or formatting purposes and does \emph{not} impact the core methodology, scientific rigor, or originality of the research, declaration is not required.
    \item[] Answer: \answerNA{} 
    \item[] Justification: LLMs are not used as an important, original, or non-standard component of the proposed method, experiments, or scientific claims. Any use of LLMs, if any, is limited to writing, editing, or formatting assistance and does not affect the core methodology or results.


\end{enumerate}